\documentclass[a4paper,fleqn]{cas-dc}

\usepackage[numbers]{natbib}
\usepackage{array}

\def\tsc#1{\csdef{#1}{\textsc{\lowercase{#1}}\xspace}}
\tsc{WGM}
\tsc{QE}
\tsc{EP}
\tsc{PMS}
\tsc{BEC}
\tsc{DE}

\DeclareMathOperator*{\argmax}{arg\,max}

\usepackage[ruled,vlined]{algorithm2e}
\hypersetup{
  colorlinks=False,
  citecolor=Violet,
  linkcolor=Red,
  urlcolor=Blue}

\usepackage{tabularx}
\usepackage{multirow}
\usepackage{makecell}
\usepackage{subcaption}
\usepackage{utfsym}
\usepackage{xcolor}
\usepackage{soul}
\usepackage{amsthm} 
\usepackage{amsmath}
\newtheorem{assumption}{Assumption}[section] 
\usepackage[switch]{lineno}

\begin{document}
\emergencystretch 3em 
\let\WriteBookmarks\relax
\def\floatpagepagefraction{1}
\def\textpagefraction{.001}
\shorttitle{Safe reinforcement learning with online filtering for fatigue-predictive human-robot task planning and allocation in production}
\shortauthors{Jintao Xue et~al.}

\title [mode = title]{Safe reinforcement learning with online filtering for fatigue-predictive human-robot task planning and allocation in production}                 



\author[1]{Jintao Xue}[]
\author[1]{Xiao Li}[orcid=0000-0001-9702-4153]
\cormark[1]
\author[1]{Nianmin Zhang}[]
\affiliation[1]{organization={CI3 Lab, Department of Civil Engineering, The University of Hong Kong},
                city={Hong Kong SAR},
                }

\cortext[cor1]{Corresponding author: CI3 Lab, Department of Civil Engineering, Faculty of Engineering, HW6-07, Haking Wong Building, The University of Hong Kong, Pokfulam, Hong Kong, China.}

\cortext[0]{This is the accepted manuscript of an article accepted for publication in
\textit{Journal of Manufacturing Systems (Elsevier)}.
The final published version is available at
\url{https://doi.org/10.1016/j.jmsy.2025.12.019}. Code under development for open release: \url{https://github.com/jintaoXue/Isaac-Production}.}

\begin{abstract}
Human-robot collaborative manufacturing, a core aspect of Industry 5.0, emphasizes ergonomics to enhance worker well-being. This paper addresses the dynamic human-robot task planning and allocation (HRTPA) problem, which involves determining when to perform tasks and who should execute them to maximize efficiency while ensuring workers’ physical fatigue remains within safe limits. The inclusion of fatigue constraints, combined with production dynamics, significantly increases the complexity of the HRTPA problem.
Traditional fatigue-recovery models in HRTPA often rely on static, predefined hyperparameters. However, in practice, human fatigue sensitivity varies daily due to factors such as changed work conditions and insufficient sleep. To better capture this uncertainty, we treat fatigue-related parameters as inaccurate and estimate them online based on observed fatigue progression during production.
To address these challenges, we propose PF-CD3Q, a safe reinforcement learning (safe RL) approach that integrates the \textbf{p}article \textbf{f}ilter with \textbf{c}onstrained \textbf{d}ueling \textbf{d}ouble \textbf{d}eep \textbf{Q}-learning for real-time fatigue-predictive HRTPA. Specifically, we first develop PF-based estimators to track human fatigue and update fatigue model parameters in real-time. These estimators are then integrated into CD3Q by making task-level fatigue predictions during decision-making and excluding tasks that exceed fatigue limits, thereby constraining the action space and formulating the problem as a constrained Markov decision process (CMDP).
Experimental results demonstrate that our PF-based estimators achieve high prediction accuracy and strong noise robustness, and that PF-CD3Q outperforms other algorithms across multiple performance metrics, significantly reducing the occurrence of overwork and adapting to unseen fatigue constraints after training. These findings validate the effectiveness of our approach under complex and dynamic production conditions, supporting both human well-being and the development of a more sustainable and efficient manufacturing paradigm.

\end{abstract}




\maketitle

\section{Introduction} \label{sec:intro}

Industry 4.0 has enhanced production efficiency and quality by integrating advanced technologies, including artificial intelligence (AI), robotics, and information technology \cite{huang2022industry}. Industry 5.0, however, represents a paradigm shift that goes beyond automation and digitization to emphasize human-centric values, including worker well-being, creativity, and cognitive engagement. A key pillar of this vision is human-robot collaboration (HRC), where humans and intelligent robots work in synergy, leveraging their complementary strengths. However, in real-world manufacturing environments, human workers often experience fatigue, which can adversely affect human performance \cite{li2025q}. Therefore, ergonomic considerations should be integrated into the HRC process to ensure sustainable production.

Task planning and allocation (TPA) refers to the process of determining when to perform specific tasks and allocating them to suitable agents—whether humans, robots, or machines—to fulfill specific manufacturing requirements \cite{cheng2019task, lee2022task, zhao2025safety}. It plays a crucial role in enhancing the effectiveness of advanced manufacturing systems. Within the context of HRC, this challenge becomes more complex, evolving into the human-robot TPA (HRTPA) problem, where both human workers and robotic agents must be dynamically coordinated to achieve optimal system performance.

This paper addresses the fatigue-constrained HRTPA problem in a dynamic, shared production environment where workers and robots collaborate on various tasks, including material transport, loading, welding, and machine operation. Key challenges include: (1) real-time HRTPA to adapt to dynamic production; (2) physical fatigue constraints for safe human workloads; (3) varying worker efficiency due to fatigue accumulation; and (4) inaccurate hyperparameters in human fatigue models, requiring real-time estimation during production. These factors underscore the demand for intelligent, adaptive, and fatigue-predictive HRTPA strategies.
Prior research faces challenges in addressing complex and dynamic HRTPA problems \cite{merlo2023ergonomic}. Some studies rely on static modeling with fully pre-known fatigue–recovery parameters and use optimization-based algorithms, but these approaches struggle with real-time decision-making and are unable to tackle inaccuracies in fatigue models \cite{cai2023task}. Other works employ reinforcement learning (RL), which is well-suited for sequential real-time decisions, yet they either overlook ergonomic constraints altogether \cite{zhang2022reinforcement} or incorporate them only through reward shaping, which still risks violating fatigue limits \cite{liu2023integration}. Although safe RL has shown potential for handling safety-critical decision-making under constraints \cite{alshiekh2018safe}, its application to real-time, fatigue-constrained HRTPA remains unexplored.

To overcome these limitations, we present a real-time fatigue-predictive HRTPA algorithm, the \textbf{p}article \textbf{f}ilter with \textbf{c}onstrained \textbf{d}ueling \textbf{d}ouble \textbf{d}eep \textbf{Q}-learning (PF-CD3Q), aimed at enhancing production efficiency while prioritizing worker ergonomics. Unlike conventional RL methods that incorporate fatigue as soft constraints via weighted reward design, PF-CD3Q adopts a safe RL paradigm with an explicit, fatigue-constrained strategy.
Specifically, while fatigue-recovery models \cite{jaber2013incorporating, cai2023task} are widely used in HRTPA, they typically assume static, pre-known hyperparameters, ignoring the temporal variability of human physical capacity. In practice, human fatigue sensitivity is subject to daily variability influenced by factors like sleep quality \cite{dawson2005managing} and environmental conditions \cite{meegahapola2018impact}. To reflect this, we treat fatigue model hyperparameters as initially inaccurate and estimate them online using particle filters (PFs) during production.
These PF-based estimators track real-time physical fatigue evolution and are used to define a safe task-level action set by masking actions likely to violate fatigue constraints. This restricts policy learning and action selection to ergonomically safe task decisions, forming a constrained Markov decision process (CMDP). To model the complex and heterogeneous information present in the production environment, our network adopts an attention-based Transformer architecture \cite{vaswani2017attention}. This design facilitates effective processing of diverse inputs, including fatigue-related data, to enable informed and adaptive decision-making.
The action output of PF-CD3Q follows a two-step process: first, selecting the task to be executed, and second, allocating it to a human and/or robot. Both steps are guided by task-level fatigue predictions generated by PF-based estimators. In the first step, the safe RL strategy determines the real-time task selection based on fatigue-predictive task planning. In the second step, we adopt a path-planning-based, distance-greedy allocation strategy to allocate tasks to the nearest available human and/or robot. 
Real-world studies of multi-participant human-robot collaborative production lines are costly and complex. To overcome this, we developed a realistic production line in NVIDIA’s Isaac Sim physical simulator \cite{liang2018gpu}, modeled after a real-world factory, integrated with a well-defined human fatigue-recovery model for simulating fatigue dynamics.

The contributions of this work include:
(1) This study novelly implements a safe reinforcement learning method for the HRTPA problem, updating the real-time safe action space by online estimators that address dynamic, explicit fatigue-constrained decision-making challenges, and can adapt to unseen fatigue constraints after training.
(2) We employ a particle filter-based estimation mechanism to reduce uncertainties stemming from inaccurate fatigue model hyperparameters, effectively addressing the challenge of daily variability in human fatigue sensitivity, and exhibit robustness to measurement noise.
(3) We design an attention-based Transformer architecture to process heterogeneous data and incorporate fatigue-related information, enhancing our algorithm’s fatigue-predictive decision-making performance.

The paper is organized as follows: Section 2 reviews related work on HRTPA, ergonomics, safe reinforcement learning, online parameter estimation, and research gaps. Section 3 formulates the fatigue-constrained HRTPA problem. Section 4 presents our methodology, detailing the proposed PF-CD3Q algorithm and the attention-based Transformer network. Section 5 demonstrates the experimental setup and comprehensive results, validating the effectiveness of our approach through comparative analysis with baseline methods. Section 6 discusses limitations and outlines future research directions.

\section{Literature review}
This section provides a comprehensive review of related work. First, we examine ergonomics in manufacturing and fatigue-recovery models. Next, we explore human-robot task planning and allocation (HRTPA) in manufacturing, summarizing representative algorithms. We then discuss safe reinforcement learning and neural network architectures. Finally, we briefly review filtering techniques for online parameter estimation.

\subsection{Ergonomics in manufacturing} \label{sec:review_ergo}
Industry 4.0, a global paradigm for over a decade, boosts profitability through technology but often neglects environmental and social factors, making it less human-centric \cite{leng2022industry}. While integrating human-robot collaboration and assistive technologies \cite{xu2021industry}, Industry 5.0 prioritizes societal goals, fostering resilient prosperity by respecting planetary boundaries and centering worker well-being. It emphasizes ergonomics across manufacturing stages (e.g., design, production, inspection, logistics), addressing human factors like performance (e.g., fatigue, breaks, skills, learning), safety (e.g., injury risks), and perceptual and environmental influences (e.g., noise exposure, visual considerations) \cite{keshvarparast2024ergonomic, prunet2024optimization}.

Human factors are critical in HRTPA and real-time operations. Among these, fatigue is a key factor affecting human well-being and can be primarily categorized into physical and psychological fatigue. Psychological fatigue, stemming from boredom or the cognitive demands of repetitive or complex tasks, may reduce production efficiency \cite{faccio2023human}. Physical fatigue has garnered more attention in the literature due to its direct impact on production efficiency, particularly in scenarios involving high physical workloads and cognitively simple tasks. Furthermore, several models have been proposed to quantify physical fatigue, assess their practical relevance, and address the complexity of integrating them into mathematical optimization frameworks \cite{prunet2024optimization}.
For example, work time is divided into active and recovery periods, with physical fatigue increasing exponentially over time and recovery diminishing similarly \cite{konz1998work}. A fatigue-recovery model with exponential dynamics accounts for varying maximum endurance times across production batches due to fatigue accumulation. Additionally, learning-forgetting dynamics, which affect production efficiency based on task familiarity, are considered \cite{ostermeier2020impact, jaber2013incorporating}. Fatigue coefficients also influence task completion times \cite{digiesi2009effect}. Analytical models have been proposed to optimize operator recovery time for better TPA \cite{calzavara2019model}, while fatigue-recovery models have been applied to HRTPA in collaborative assembly cells \cite{cai2023task}. Asadayoobi et al. \cite{asadayoobi2023optimising} tackle stochastic bi-objective TPA, incorporating the combined effects of learning-forgetting, fatigue-recovery, and stress-recovery processes. Similarly, Ferjani et al. \cite{ferjani2017simulation} propose a simulation-optimization heuristic for online allocation of multi-skilled workers subject to fatigue in manufacturing systems.
Liu et al. \cite{liu2023integration} combine a fatigue and recovery model with deep reinforcement learning and a multi-agent system for dynamic scheduling in re-entrant hybrid flow shops.

Based on the literature reviewed, this paper focuses on physical fatigue due to its extensive coverage and established mathematical models.
A worker’s physical fatigue sensitivity is influenced by stable static factors (e.g., skills, age, gender, task type) and daily psycho-physical conditions \cite{dawson2005managing} or environmental factors such as temperature, humidity, and noise \cite{meegahapola2018impact}. Most fatigue models in production HRTPA rely on fixed or predefined hyperparameters, overlooking daily fluctuations in these dynamic factors. To better capture real-time fatigue and productivity shifts, adaptive methods like online parameter estimation are essential. These techniques dynamically adjust model parameters based on observed worker performance, facilitating more responsive and realistic scheduling decisions.

\subsection{Task planning and allocation in human-robot collaboration}

Task planning and allocation (TPA) aims to optimize existing resource arrangements to enhance manufacturing systems' economic performance and social benefits \cite{cheng2019task}. TPA typically involves three main stages: task description, modeling, and algorithm design. In the task description stage, tasks are defined based on production processes and available resources. Timo et al. \cite{banziger2020optimizing} used standardized descriptions for mobile assistant robots, while Li et al. \cite{li2023knowledge} proposed semantically enriched packages to enhance task understanding. In the modeling stage, formal approaches like mixed-integer programming (MIP) \cite{faccio2024task} are used for deterministic problems, whereas some works use Markov decision processes (MDP) to model sequential decision-making under uncertainty. The algorithm design stage includes both traditional optimization methods, such as genetic algorithms for TPA in assembly \cite{patel2020decentralized} and hybrid strategies for multi-agent systems \cite{fontes2023hybrid}, and learning-based techniques, including reinforcement learning (RL), which enables real-time feedback and adaptive decisions \cite{lee2022digital}.

Human-robot collaboration (HRC) refers to humans and robots working together in a shared workspace toward common goals \cite{mukherjee2022survey}. With advancements in automation and information technology, robots increasingly handle repetitive, high-precision, or physically demanding tasks, allowing humans to focus on cognitive and flexible operations. This complementarity has made HRC a core research focus in intelligent manufacturing \cite{dhanda2025reviewing}. HRC applications span assembly, welding, assistive operations, logistics, and quality inspection \cite{kim2020estimating, lu2023human, peternel2019selective, matheson2019human}, with TPA optimizing efficient coordination between diverse robots and humans. HRC configurations include collaborative robots (cobots), robotic arms, mobile robots, exoskeletons, and specialized production robots \cite{gong2022toward, krupas2023human, malik2024intelligent}.

Research on HRC spans multiple domains, including ergonomic work cell layout design \cite{cherubini2016collaborative, wang2025design}, human behavior modeling, perception-based interaction \cite{wang2018deep, hietanen2020ar, liu2021brainwave, zheng2023video, zheng2025human, wang2025deep, zheng2025human}, and TPA. A key focus within this field is human-robot TPA (HRTPA), which aims to optimize collaboration by leveraging the complementary capabilities of humans and robots. Given that manufacturing tasks often involve sequential subtasks, effective allocation must consider factors such as cost, makespan, and ergonomics. 
To address this, various approaches have been proposed: (1) Search-based methods find the solution in a finite and predefined state space \cite{asadayoobi2023optimising}. For instance, Merlo et al. \cite{merlo2023ergonomic} developed a framework incorporating ergonomic factors into dynamic TPA, while Zeng et al. \cite{zeng2025task} applied multi-heuristic local search and fast greedy refinement. However, these methods face scalability issues as the problem size grows. (2) Optimization-based approaches aim to formulate HRTPA problems mathematically \cite{zhao2025safety}. Yao et al. \cite{yao2024task} proposed a genetic algorithm enhanced with reinforcement learning and a dynamic fatigue model, while Cai et al. \cite{cai2023task} introduced a multi-objective MIP model considering both physical and psychosocial fatigue. These methods, though effective in structured settings, often struggle with real-time adaptability and the modeling of dynamic factors such as the spatial position of humans and robots in the workspace or fluctuating human fatigue and production efficiency. (3) RL offers a promising solution for sequential decision-making under uncertainty \cite{xue2026hierarchical}. Zhang et al. \cite{zhang2022reinforcement} proposed an RL-based method for task sequence optimization in assembly, and Liu et al. \cite{liu2023integration} integrated RL for adaptive HRC. However, existing RL approaches often treat ergonomics as a soft constraint through the reward function, which may still result in fatigue violations and compromise worker safety. 

\subsection{Safe reinforcement learning}
Reinforcement learning (RL) refers to the agent maximizing long-term returns by interacting with an environment through state observations, actions, and reward signals \cite{sutton1998reinforcement}.
However, in safety-critical applications such as autonomous driving or industrial automation, ensuring the safety of actions is paramount to prevent damage or catastrophic failures. Safe RL extends vanilla RL by incorporating safety constraints during learning and/or deployment, often modeled as a constrained Markov decision process (CMDP) \cite{altman1993asymptotic}. CMDPs extend MDPs by incorporating constraints alongside rewards, enabling the optimal policy to maximize cumulative returns while ensuring expected costs remain within specified limits \cite{altman2021constrained}.

Safe RL methods can be broadly categorized into constraint optimization-based and knowledge-utilization approaches \cite{gu2024review}. 
Optimization-based methods enforce safety by incorporating cost constraints into policy optimization \cite{chow2019lyapunov, achiam2017constrained, ray2019benchmarking, stooke2020responsive, zhang2022penalized, liu2020ipo, tessler2018reward}. Constrained policy optimization (CPO) \cite{achiam2017constrained} updates the policy within a trust-region constraint and performs a conjugate-gradient–based optimization step, thereby guaranteeing theoretically bounded policy improvement. However, this second-order mechanism is computationally expensive and highly sensitive to training noise in non-convex environments, leading to instability and eventual divergence during training. Lagrangian methods \cite{ray2019benchmarking, stooke2020responsive} convert constrained problems into unconstrained ones using Lagrange multipliers, though they can be unstable. To improve this, Stooke et al. \cite{stooke2020responsive} introduced a PID Lagrangian approach, enhancing stability through control-theoretic adjustments. Penalty-based methods like IPO \cite{liu2020ipo} add barrier functions to reduce violations. However, the above optimization-based methods may still violate constraints in the training and deployment stage, achieving approximate constraint satisfaction \cite{gu2024review}.

Knowledge-utilization methods enhance safety by incorporating prior knowledge to guide exploration, avoiding risky actions more directly than constraint optimization-based approaches \cite{geramifard2013intelligent, moldovan2012safe, kalweit2020deep, xu2022constraints, alshiekh2018safe}.
Moldovan et al. \cite{moldovan2012safe} propose a safe exploration framework that optimizes within guaranteed safe policies, ensuring ergodicity with user-specified probability ($\delta$-safe policies), ideal for non-ergodic physical systems. 
Alshiekh et al. \cite{alshiekh2018safe} enforce safety specifications in reinforcement learning by synthesizing a “shield” that either filters or corrects the agent’s actions, ensuring safe policy learning while preserving convergence guarantees.
Some are integrated with the Q-learning-based RL paradigm, selecting actions via the argmax of value functions without requiring an explicitly defined policy \cite{ji2024omnisafe}.
Kalweit et al. \cite{kalweit2020deep} propose deep constrained Q-learning, demonstrating that safety can be ensured in discrete policy extraction by restricting the action space during action-value maximization. Similarly, in offline RL, Xu et al. \cite{xu2022constraints} propose constraints penalized q-learning, modifying the Bellman update to penalize unsafe state-action pairs, including those outside the data distribution. Knowledge-based methods in safe RL, well-suited for Q-learning frameworks, excel in the HRTPA problem in manufacturing, where discrete actions like allocating tasks to robots or humans are common \cite{cheng2019task}. Overall, safe RL holds strong potential for enhancing worker safety, well-being, and operational efficiency in production environments.

 \subsection{Filters for online parameter estimation}
 
Online state and parameter estimation is essential in fields such as signal processing, robotics, and control systems, where systems evolve dynamically, and data arrive sequentially \cite{chen2011kalman, gustafsson2002particle}. In such settings, some system parameters or states are either unknown or affected by measurement noise, requiring techniques to filter observations or estimate hidden variables.

The Kalman filter (KF) provides optimal state estimation for linear systems with Gaussian noise, using system inputs and outputs to reduce noise effects recursively \cite{welch1995introduction}. For nonlinear systems, the extended Kalman filter (EKF) applies linear approximations, while the unscented Kalman filter (UKF) uses unscented transformations for higher accuracy \cite{chen2003bayesian}. Particle filter (PF), based on Monte Carlo methods, approximates posterior distributions using weighted particles, making them suitable for highly nonlinear, non-Gaussian tasks such as SLAM or fault diagnosis \cite{ gustafsson2002particle}. Despite their flexibility, PFs face challenges like particle degeneracy and high computational cost in high-dimensional spaces.

In HRTPA for production, human fatigue-recovery models are widely used. However, most approaches assume static or predefined hyperparameters, overlooking daily variability in human performance (see Section \ref{sec:review_ergo}). To overcome this, this paper treats the model's hyperparameters as initially inaccurate and employs real-time filtering to estimate them online, enabling more adaptive and responsive HRTPA in dynamic production environments.

\subsection{Research gaps}
In summary, the above literature review focuses on HRTPA in dynamic production environments, ergonomics, safe RL, and online parameter estimation. However, current research has the following limitations:

(1) Current HRTPA research struggles to address inaccurate or unknown fatigue model parameters, typically assuming these coefficients are predefined and known a priori, while overlooking the inherent daily variability in human fatigue sensitivity and individual differences in fatigue accumulation patterns.

(2) Despite the progress of RL-based approaches for the HRTPA problem, they often neglect fatigue factors or face constraint violations when solving constrained problems, leading to excessive workload for humans in production.

(3) While safe RL provides a principled framework for constraint-aware decision-making, prior works are rarely applied to HRTPA and seldom consider explicit ergonomic constraints or incorporate fatigue dynamics into the safety mechanism.

\section{Problem Formulation}\label{sec:problem}
\begin{figure*}[htb] 
\centering	
	\includegraphics[width=0.99 \linewidth, height=0.6\linewidth]{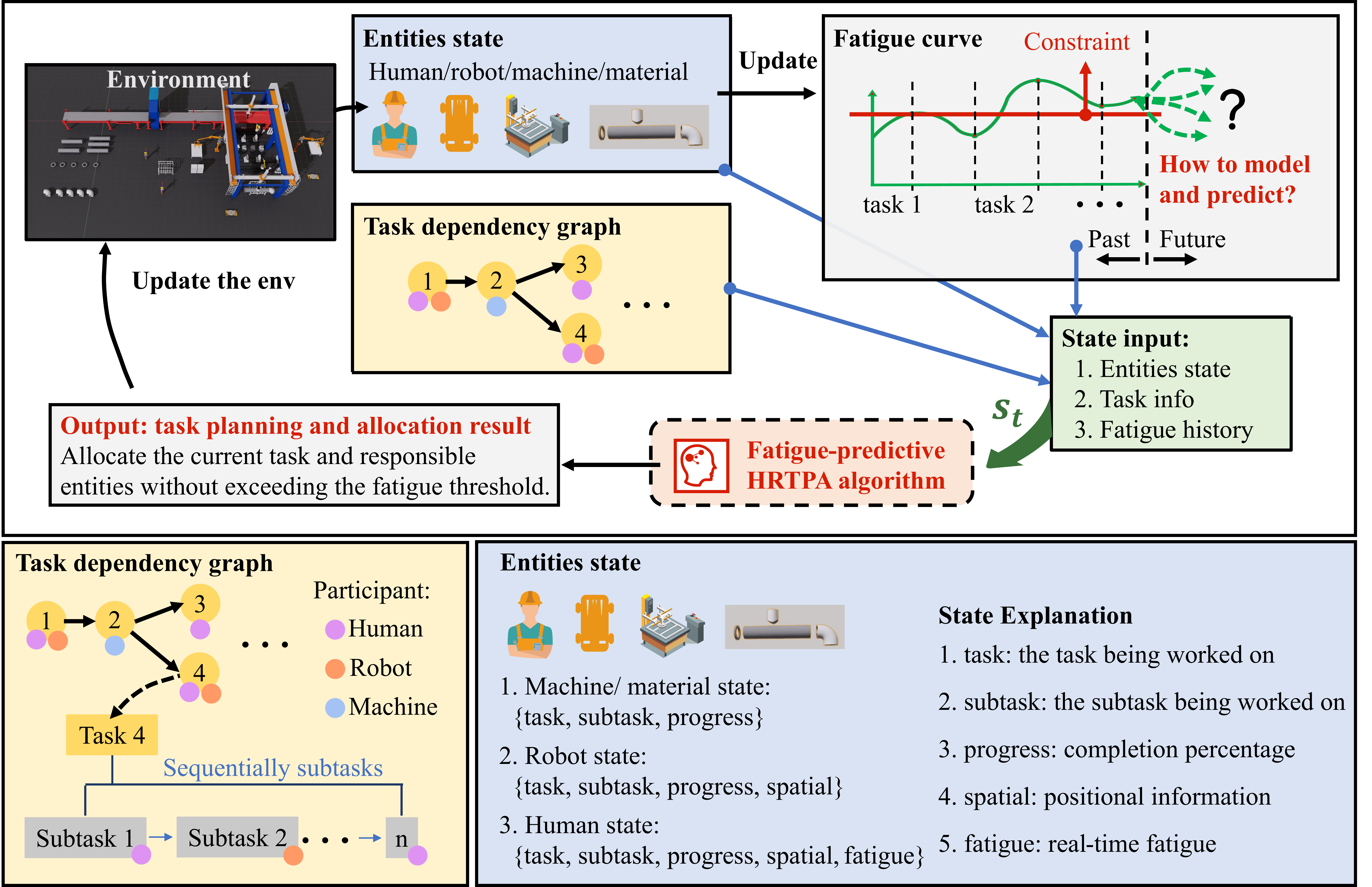}
\caption{Illustration of the real-time production process flow and state information. \label{fig:problem}} 
\end{figure*}

We present a description and formulation of the human-robot collaborative production problem. Figure \ref{fig:problem} illustrates the process flow, state information, task dependency graph, and fatigue constraints, clarifying the formulation. We then detail task decomposition, entity states, the real-time fatigue model with its uncertain parameters for our HRTPA algorithm (see Assumption \ref{ass:parameter}), dynamic task completion efficiency, and the objective function.

\subsection{Task decomposition}\label{sec:Task}
Following prior work \cite{cheng2019task}, we predefine production tasks and their manageable subtasks in alignment with the nature of the production line. The dependency graph \(\mathcal{T}\) and task decomposition are expressed as follows:
\begin{equation} \label{equation:task}
\begin{gathered}
    \mathcal{T} = \{\mathcal{T}^{h}, \mathcal{T}^{r}, \mathcal{T}^{m}\} = \{task_0, task_1, ..., task_i, ...\}, \\
    \mathcal{O} = \{ \mathcal{O}^{h}, \mathcal{O}^{r}, \mathcal{O}^{m} \} =\{subtask_0, ..., subtask_i, ...\}, \\ 
   task_i = \{subtask_{i,0}, subtask_{i,1}, \ldots, subtask_{i,j}, \ldots\},\\
   task_i \in \mathcal{T}, subtask_{i,j} \in \mathcal{O},
\end{gathered}
\end{equation}
where \(\mathcal{T}\) defines tasks for humans, robots, and machines, with subsets \(\mathcal{T}^{h}\), \(\mathcal{T}^{r}\), and \(\mathcal{T}^{m}\) representing tasks for each entity, respectively. \(\mathcal{O}\) encompasses all subtasks, with \(\mathcal{O}^{h}\), \(\mathcal{O}^{r}\), and \(\mathcal{O}^{m}\). Each task \(task_i\) comprises a predefined, sequential subtask sequence tailored to the production problem, providing detailed, step-by-step instructions for humans, robots, and machines. 

\subsection{Entities state description}\label{sec:entity}
Consider a production line involving groups of operator entities, including robots, humans, and machines, defined as follows:
\begin{equation} \label{equation:entity}
\begin{gathered}
    {\mathcal{E} = \{\mathcal{E}^h, \mathcal{E}^r, \mathcal{E}^{mac}, \mathcal{E}^{mat}\}},
\end{gathered}
\end{equation}
where \(\mathcal{E}\) denotes the set of all operator entities. The robot set is defined as \(\mathcal{E}^r = \{e^r_0, e^r_1, ..., e^r_i, ...\}\), where \(e^r_i\) represents the \(i\)-th robot. Similarly, \(\mathcal{E}^h\) denotes the human set, \(\mathcal{E}^{mac}\) the machine set, and \(\mathcal{E}^{mat}\) the raw materials set. The entity states are defined as follows:
\begin{equation} \label{equation:task_subtask}
\begin{gathered}
   \mathbf{s}^h_{k,t} = \{task, subtask, progress, spatial, fatigue\}, \\
   \mathbf{s}^r_{i,t} = \{task, subtask, progress, spatial\}, \\
    \mathbf{s}^{mac}_{j,t}, or \; \mathbf{s}^{mat}_{m,t}= \{task, subtask, progress\}, \\
    where, \; spatial=\{x,y\}, fatigue=F_{k,t},\\
 e^h_k, e^r_i, e^{mac}_j,  e^{mat}_m\in \mathcal{E}, task \in \mathcal{T}, subtask \in \mathcal{O}, \\
\end{gathered}
\end{equation}
where, at each time step $ t $, \( task \in \mathcal{T} \) and \( subtask \in \mathcal{O} \) denote the current task and subtask allocated to the human, robot, or machine. The \( progress \) variable reflects the task completion degree based on its nature, while \( spatial = \{x, y\} \) provides real-time global position information for movable humans or robots. Notably, the human fatigue state, a critical factor, is also considered. Following prior work \cite{liu2023integration, cai2023task}, we employ a fatigue-recovery model to simulate human fatigue changes during production, with further details provided in the later section.

\subsection{Environment setup for human fatigue and task completion time}\label{sec:fatigue}

To reflect the dynamic nature of the production process, during the problem formulation stage, we introduce the modeling of human physical fatigue and its effect on varying task processing efficiency.
We then demonstrate how task completion time is calculated under real-time efficiency changes.

In the context of work, time is divided into working and resting periods. Fatigue accumulates exponentially during work, while the effectiveness of rest decreases exponentially over time, as noted in prior research \cite{konz1998work}. To simulate worker fatigue accurately, we adopt the fatigue-recovery exponential model from earlier studies \cite{jaber2013incorporating}. The mathematical formulations are presented in Eq. \ref{eq:fatigue}.

\begin{equation} \label{eq:fatigue}
\begin{gathered}
F_{k,t}=
\begin{cases} 
    F_{k,t-1} e^{-\mu_k}, \quad\quad\quad\; \text{if resting}, \\
    F_{k,t-1} + (1 - F_{k,t-1})(1 - e^{-\lambda_{k,ij}}),\\
    \quad\quad\quad\quad\quad\quad\quad\quad \text{if doing } subtask_{i,j}.\\
\end{cases} \\
    e^h_k \in \mathcal{E}^h, subtask_{i,j} \in task_i,
    \end{gathered}
\end{equation}
where \( F_{k,t} \) represents the real-time fatigue of the \( k \)-th human at time \( t \), \( e^h_k \in \mathcal{E}^h \). The parameter \( \mu_k \) influences the recovery speed, while \( \lambda_{k,ij} \) affects the fatigue accumulation rate, varying with each \( subtask_{i,j} \in task_i \). During the rest stage, the human is assumed to remain at their previous location without moving, waiting for the next task allocation. Once a new task is received, the human transitions to the working state. If no task is allocated, the human’s fatigue state is updated and naturally decreases according to the fatigue recovery model described above. Since production performance may vary daily for the same human, potentially impacting recovery (\( \mu_k \)) or fatigue accumulation rates (\( \lambda_{k,ij} \)), we adopt the following assumption:

\begin{assumption} \label{ass:parameter}
We assume that the hyperparameters of the human fatigue model are initially inaccurate for the HRTPA algorithm and require real-time estimation during the production process.
\end{assumption}

The above assumption poses challenges for fatigue-predictive HRTPA, as the fatigue model parameters are initially partially unknown for the algorithm. We introduce uncertainty by initializing the parameters from the following Gaussian distribution noise:

\begin{equation} \label{eq:fatigue_noise_param}
\begin{gathered}
\lambda_{init} = \lambda_{true}\cdot(1+randomness),\\
\mu_{init} = \mu_{true}\cdot(1+randomness),\\
randomness \sim \mathcal{N}(0, \sigma_{init}),
\end{gathered}
\end{equation}
where $\lambda_{init}$ and $\mu_{init}$ are the initial parameter values known to the algorithm, and $\sigma_{init}$ controls the degree of initial parameter uncertainty.

In the simulation, human fatigue evolves according to Eq. \ref{eq:fatigue} using the ground-truth parameters ($\mu, \lambda$). In contrast, the online algorithm has access only to fatigue measurements and lacks knowledge of the true model parameters according to Eq. \ref{eq:fatigue_noise_param}, thereby requiring online parameter estimation to improve long-term, task-level fatigue prediction.
We describe how our approach addresses this challenge in Sec. \ref{sec:method}. Meanwhile, worker production efficiency is influenced by fatigue, as noted in prior work \cite{liu2023integration, jaber2013incorporating}, as follows:

\begin{equation} \label{eq:efficiency}
\begin{gathered}
\tau_{k,i,j,t} = \tau_{i,j}(1+\delta_{eff}(ln(1+F_{k,t}))),\\
efficiency_{k,i,j,t} = 1/{\tau_{k,i,j,t}}, \\
e^h_k \in \mathcal{E}^h, subtask_{i,j} \in task_i,\\
\end{gathered}
\end{equation}
in which \(\tau_{i,j}\) represents the predefined static completion time of \(subtask_{i,j}\), and \(\delta_{eff}\) is a hyperparameter controlling the time change scale related to human fatigue. However, the computation of the completion time for the k-th human performing \(subtask_{i,j}\), denoted \(\tau_{k,i,j,t}\) assumes static fatigue during the execution of \(subtask_{i,j}\), which conflicts with the real-time fatigue variation setting. Thus, we compute the real-time efficiency \(efficiency_{k,i,j,t}\) for \(subtask_{i,j}\), reflecting the completion degree per time step. Human fatigue and production efficiency are updated in real-time, increasing the completion degree (0 for start, 1 for finished) by summing \(efficiency_{k,i,j,t}\) multiplied by unit time until reaching finished.
Through the above formulas and Assumption \ref{ass:parameter}, we account for the impact of fatigue change on real-time production efficiency and the uncertainty in task completion time.

Eq. \ref{eq:efficiency} shows fatigue accumulation directly influences real-time production efficiency by inducing variability in human performance, which in turn affects subtask completion times ($subtime_{i,j}$). To further account for stochastic fluctuations in subtask completion time during the production process, we introduce additional randomness terms into the $subtime_{i,j}$. Specifically, the actual execution time of $subtask_{i,j}$ is modeled as:
\begin{equation} \label{eq:task_time_random}
\begin{gathered}
subtime_{random,i,j} = \\subtime_{i,j}\cdot(1+randomness) + t_{travel,i,j},\\
randomness \sim \mathcal{N}(0, \sigma_{time}),
\end{gathered}
\end{equation}
where $subtime_{random,i,j}$ captures processing-time variability (e.g., minor disturbances, traveling time), $\sigma_{time}$ is the hyperparameter. $t_{travel,i,j}$ represents the travel time spent for the human and/or robot to reach and return for the subtask, which is challenging to predict deterministically.

\begin{figure*}[htb] 
\centering	
	\includegraphics[width=0.99 \linewidth, height=0.7\linewidth]{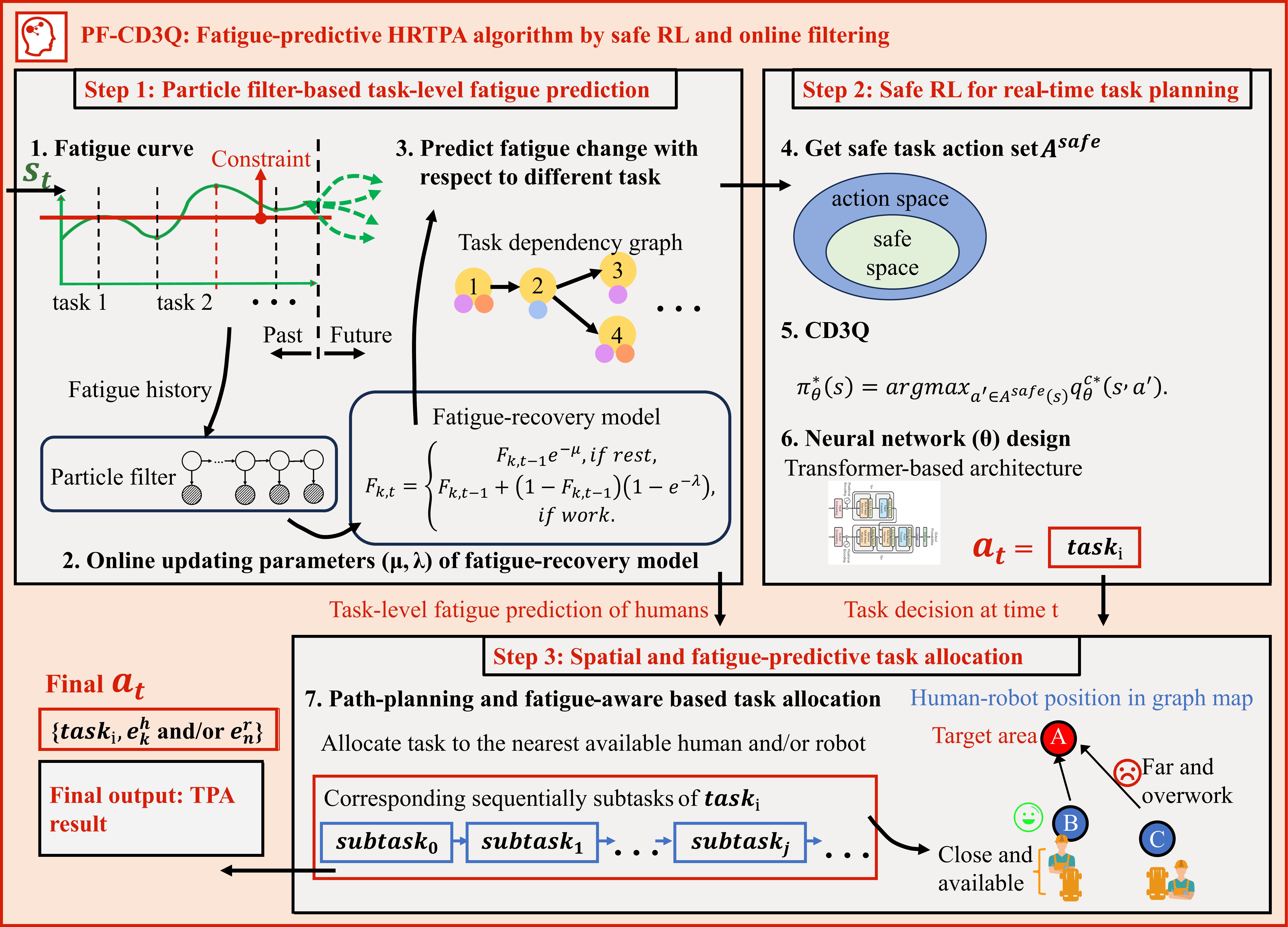}
\caption{Illustration of our method PF-CD3Q: fatigue-predictive HRTPA algorithm with CD3Q and online filtering. \label{fig:algorithm}} 
\end{figure*}

\subsection{Real-time state and objective function}\label{sec:obj}
The real-time state of the production process at time step \( t \) is defined as:
\begin{equation} \label{eq:state}
\begin{gathered}
    \mathbf{s}_{t} = \{\mathbf{s}^h_t, \mathbf{s}^r_t, \mathbf{s}^{mac}_t, \mathbf{s}^{mat}_t, \mathcal{T}, \mathcal{O}, Grid\_Map\},
\end{gathered}
\end{equation}
where state \(\mathbf{s}_{t}\) encompasses the states of available operator entities (humans, robots, machines, as defined in Sec. \ref{sec:entity}) and task-related information (Sec. \ref{sec:Task}). The \(Grid\_Map\) is an occupancy grid map derived from the raw environment, classifying each grid cell as free space or occupied (indicating obstacles).
We aim to design an HRTPA algorithm, taking state \(\mathbf{s}_t\) as input and producing a result while adhering to fatigue constraints:
\begin{equation} \label{eq:obj_1}
\begin{gathered}
   a_t = \mathrm{HRTPA}(s_t),\\
    where \; a_t = \{task_i, e_m^h \; and/or \; e_n^r \},\\
    s.t. \; \forall \; F_{k,t}, F_{k,t} < d_{k}, \\
    task_i \in \mathcal{T}, e_k^h, e_m^h \in \mathcal{E}^h, e_n^r \in \mathcal{E}^r,
\end{gathered}
\end{equation}
where \(a_t\) specifies the current \(task_i\), its sequential subtasks, and the allocated human and/or robot entities \(e_k^h, e_m^r\). The action \(a_t\) should be selected to ensure that the fatigue threshold \(d_k\) for any human is not exceeded.
Subsequently, \(a_t\) guides each entity, particularly humans and robots, to perform suitable tasks and subtasks at each time step until completion, defined as:
\begin{equation} \label{eq:obj_2}
\begin{gathered}
\begin{cases} 
    s_{t+1} = Env(a_t), \text{if not end}, \\
    makespan = t,  \text{if end}.\\
\end{cases} \\
    \end{gathered}
\end{equation}

The production environment \(Env\) updates based on \(a_t\), generating a sequence of allocations: \(a_0, a_1, \ldots, a_t\). The objective is to minimize the total production time, denoted as \(makespan\), while respecting fatigue constraints.

\section{Research method} \label{sec:method}
\begin{algorithm*}[htb]
\small \caption{PF-CD3Q: fatigue-predictive HRTPA algorithm by safe RL and online filtering.}\label{alg:PF-CD3Q}
\SetKwInOut{Input}{input}\SetKwInOut{Output}{output}
\Input{$\mathcal{D}$ -- empty prioritized replay buffer; $\theta$ -- behavior network, $\overline{\theta}$ -- target network;}
\Input{$N_r$ -- replay buffer maximum size; $N_b$ -- training batch size; $\overline{N}$ -- target network replacement freq;}
\Input{$N_{train}$ -- start training step; $N_{p}$-- buffer replay period;}
\Input{$H$ -- time horizon of one episode, $M$ -- episodes maximum size, $t_{total}$ total steps = 0;}
\For{$\textit{episode} \in \{ 1, 2, \ldots, M \}$} {
  time step t = 0;
  \While{\textit{episode} \textup{is not end}} {
    Task planning: observe $s_t$ and choose action $a_t \sim \pi_{\theta}, a_t = \{task_i\} $\tcp*{State and task description in Sec. \ref{sec:problem} }
    Task allocation: expand action \( a_t = \{task_i, e_k^h \text{ and/or } e_n^r\} \) in Sec. \ref{sec:expand_action};\\
    Sample next state $s_{t+1}$ from environment given $(s_t,a_t)$ and receive reward $r_{t+1}$;\\
    Update parameters of the human fatigue model, generate the safe action set \( A^{safe} \) and the safe human set $E^{safe}$ using Algorithm \ref{alg:PF_predict}; \\
    
    Add transition tuple $(s_t, a_t, s_{t+1}, r_{t+1})$ to $\mathcal{D}'$, set $t \leftarrow t+1$ \;
    {\bf if} $t_{total} < N_{train}$ and $(t_{total} \mod (N_{p}-1)) \not\equiv 0 $ {\bf: continue}; \\
    \For{$i \in \{0, 1, \ldots,  N_{p}-1\}$}{
        Sample a minibatch of \( N_b \) tuples \((s_t, a_t, s_{t+1}, r_{t+1}) \sim \mathcal{D}\), \\
        Update behavior work $\theta$ by doing a gradient descent step with loss: \\
        ~~~~ $y_t =  r_{t+1} + \gamma q^{c}_{\theta^-}(s_{t+1}, \argmax_{a'\in A^{safe}(s_{t+1})} q^{c}_{\theta}(s_{t+1}, a')), $ \\
        ~~~~ $L(\theta) = \mathbb{E}_{(s_t, a_t, s_{t+1}, r_{t+1}) \sim \mathcal{D}} \left[ \left( \alpha (y_t - q^c_\theta(s_t, a_t)) \right)^2 \right],$ \tcp*{Loss function in Sec. \ref{sec:rl_obj_loss} }
        Replace target network $\overline{\theta} \leftarrow \theta$ every $\overline{N}$ steps\;
    }
    } 
}
\end{algorithm*}
To address challenges in Sec. \ref{sec:intro} and the problem in Sec. \ref{sec:problem}, this section presents PF-CD3Q, combining constrained dueling double deep Q-learning (CD3Q) with the particle filter (PF) for fatigue prediction, as depicted in Fig. \ref{fig:algorithm}. First, the filter tackles Assumption \ref{ass:parameter} by estimating task-level fatigue changes online. Second, using these predictions, it generates a safe task-level action set, which outputs task planning decision ($task_i \in \mathcal{T}$) prioritizing the task to be done at time step \( t \). After task output and fatigue prediction, the algorithm employs path-planning-based task allocation, allocating the nearest fatigue-acceptable available human and/or robot (\( e_m^h \) and/or \( e_n^r \), where \( e_m^h \in \mathcal{E}^h \), \( e_n^r \in \mathcal{E}^r \)) to execute the task's predefined sequential subtasks.
In this section, we detail our PF-CD3Q algorithm, covering the particle filter, safe RL formulation, and a transformer architecture for handling heterogeneous state information.

\subsection{Particle filter-based task-level fatigue prediction} \label{sec:method_pf_predict}

Assumption \ref{ass:parameter} highlights the challenge of initially inaccurate hyperparameters in the human fatigue model, requiring real-time estimation during production. To address this, we develop particle filter-based online parameter estimators, detailed in Algorithm \ref{alg:PF_predict}, it estimates and updates parameters \(\mu_{k,t}\) and \(\lambda_{k,i,j,t}\) in real-time for the fatigue and recovery model, where \(\mu_{k,t}\) is the recovery coefficient for the \(k\)-th human (\(e_k^h \in \mathcal{E}^h\)), and \(\lambda_{k,i,j,t}\) is the coefficient for the \(k\)-th human performing \(subtask_{i,j} \in task_i\). The process starts by initializing particles \(\{\mu_{k,n,t}\}_{n=1}^{N_p}\) or \(\{\lambda_{k,i,j,n,t}\}_{n=1}^{N_p}\) and weights \(\{w_{k,n,t} = 1 / N_p\}_{n=1}^{N_p}\) (for recovery) or \(\{w'_{k,n,i,j,t} = 1 / N_p\}_{n=1}^{N_p}\) (for \(subtask_{i,j}\)). 
Specifically, we first initialize the inaccurate parameters \((\lambda_{\text{initial},k,i,j}, \mu_{\text{initial},k})\) by injecting random noise into their ground-truth values following Eq. \ref{eq:fatigue_noise_param}. Next, for each parameter, we generate an initial set of particles within a symmetric percentage range around its value. The upper and lower bounds for sampling are defined as:

\begin{equation} \label{eq:particle_initial}
\begin{gathered}
   upper\_bound = param\cdot(1+\sigma_{particle}),\\ 
   lower\_bound = param\cdot(1+\sigma_{particle}),\\
   \text{Initialize particles}: \{\mu_{k, n, t}\}_{n=1}^{N_p}, \;or\; 
     \{\lambda_{k,i,j,n, t}\}_{n=1}^{N_p} \\ \sim \mathcal{U}(low\_bound, upper\_bound),\\
     e^h_k \in \mathcal{E}^h, subtask_{i,j} \in task_i, param \; \in \;\{\mu_k, \lambda_{k,i,j}\},
\end{gathered}
\end{equation}
where each noise-initialized parameter corresponds either to $\mu_{\text{initial},k}$ for the k-th human or to $\lambda_{\text{initial},k,i,j}$ for the k-th human performing $subtask_{i,j}$ within $task_i$. The resulting $upper\_bound$ and $lower\_bound$ define a symmetric interval centered on the parameter value, from which particles are uniformly sampled. Here, $N_p$ denotes the number of particles per parameter, and $\sigma_{particle}$ is the global hyperparameter.

As production progresses, we receive ground-truth fatigue values with added Gaussian measurement noise: $z_{k,t} \sim \mathcal{N}(F_{k,t}, \sigma_m^2)$. The filters use fatigue measurements to update particle weights via a likelihood function, which are subsequently normalized and resampled when the effective sample size falls below a predefined threshold.

\begin{algorithm}[htb]
\small \caption{Subtask completion time considering fatigue for human $e_k^h \in \mathcal{E}^h$}\label{alg:subtime}
\SetKwInOut{Input}{input}\SetKwInOut{Output}{output}
\Input{$\tau_{i,j}$ -- static completion time of $subtask_{i,j}$;} 
\Input{$t$ -- current time step t;} 
\Input{$F_{k,t}$ -- current fatigue for human $e^h_k \in \mathcal{E}^h$;} 
\Input{$\mu_{k,t}, \lambda_{k,i,j,t}$ -- estimated parameters for fatigue model;} 
\Output{$subtime_{k,i,j,t}$ -- subtask completion time;}
\Output{$F_{k,t+subtime_{k,i,j,t}}$ -- subtask-level fatigue, excluding travel and stochastic fluctuations detailed in Eq.\ref{eq:task_time_random};}
Initialize subprogress = 0\tcp*{subtask completion degree}  
Initialize $subtime_{k,i,j,t} = 0$
\While{subprogress < $\mathrm{1}$}{
    $\%$ $subtask_{i,j}$ remains unfinished.\\
    Update fatigue $F_{k,t}$ using $\;$ Eq. \ref{eq:fatigue} and $\mu_{k,t}, \lambda_{k,i,j,t}$; \\
    Update $efficiency_{k,i,j,t}$ using $\;$ Eq. \ref{eq:efficiency}; \\
    subprogress += $efficiency_{k,i,j,t}$*1;\\
    t += 1, $subtime_{k,i,j,t}$+=1; \\
}
\end{algorithm}

\begin{algorithm*}[htb]

\small \caption{Particle filter-based task-level fatigue prediction.}\label{alg:PF_predict}
\SetKwInOut{Input}{input}\SetKwInOut{Output}{output}
\Input{Task information: $task_i \in \mathcal{T}^h$, $subtask_{i,j}\in task_i$ -- See Sec. \ref{sec:Task} for more task details} 
\Input{\(N_p\) -- number of particles; \(\sigma_m\) -- measurement noise }
\Output{$\mu_{k,t}, \lambda_{k,i,j,t}$ -- estimated parameters for recovery and $subtask_{i,j}$, See Sec. \ref{sec:fatigue} for fatigue model details}
\Output{$\tau_{k,i,t}$ -- completion time of $task_{i}$; $F_{k,i,t+\tau_{k,i,t}}$ -- task-level fatigue prediction}
Initialize time t = 0;\\
\While{\textit{process} \textup{is not end}} {
    $\#\#\#\#\#$ {\bf Step one: estimated parameters of fatigue model} \\
    \( F_{k,t} \) -- fatigue of the \( k \)-th human, where \( e^h_k \in \mathcal{E}^h \), fatigue is updated according to Eq. \ref{eq:fatigue} using the ground-truth values of the fatigue model parameters;\\
    Receive humans fatigue measurements $z_{k,t} \sim \mathcal{N}(F_{k,t}, \sigma_m^2)$ for $F_{k,t}$, and states information (e.g., resting or doing $subtask_{i,j}$); \\
    {\bf if} the particles of corresponding parameters $\mu_{k,t}$ or $\lambda_{k,i,j,t}$ is not initialized {\bf then:} \\
     $\quad$ Initialize uncertain parameters ($\lambda_{\text{initial},k,i,j}$, $\mu_{\text{initial}, k}$) by adding randomness to the ground-truth values according to Eq. \ref{eq:fatigue_noise_param}; \\
     $\quad$ Initialize particles \( \{\mu_{k, n, t}\}_{n=1}^{N_p}, \;or\; 
     \{\lambda_{k,i,j,n, t}\}_{n=1}^{N_p}\) are initialized by sampling from a normal distribution, as detailed in Eq. \ref{eq:particle_initial}); \\ 
    $\quad$ Initialize weights $\{w_{k, n, t}= 1 / N_p\}_{n=1}^{N_p}$ (for recovery), or $\{w'_{k,n,i,j,t} = 1 / N_p\}_{n=1}^{N_p}$ (for doing $subtask_{i,j}$), set corresponding initial prediction \(F_{\text{pred,k,t}} = z_{k,t}\); \\
    {\bf else:} \\
    $\quad$ Compute predictions $F_{\text{pred,k,n,t}}$ for each particle ($\mu_{k, n, t}$ or $\lambda_{k,i,j,n,t}$) by propagating the previous prediction $F_{\text{pred},k,t-1}$ through the fatigue dynamics (Eq.\ref{eq:fatigue}) using the estimated parameters;\\
    $\quad$ Update weights using the likelihood function: \( w_{n,t} = w_{n, t-1} \cdot \exp\left(-\frac{(z(t) - F_{\text{pred},k,n,t})^2}{2\sigma_m^2}\right) \), where \( w_{n, t} \) can be recovery weights \( w_{k,n,t} \) or subtask weights \( w'_{k,n,i,j,t} \);\\
    $\quad$ Normalize weights: \(w_{n,t} = w_{n,t} / \sum_{m=1}^{N_p} w_{m,t}\) \\ 
    $\quad$ Compute effective sample size: \(N_{\text{eff}} = 1 / \sum_{n=1}^{N_p} w_{n,t}^2\) \\
    $\quad$ if \(N_{\text{eff}} = 1\) < $N_p/2$, then: \\
    $\quad$$\quad$ Resample: systematically resample particles based on cumulative weights.\\
    Estimate fatigue parameters \(\mu_{k,t} = \sum_{n=1}^{N_p} w_{k,n,t} \cdot \mu_{k,n,t}\), \(\lambda_{k,i,j,t} = \sum_{i=n}^{N_p} w_{k,n,i,j,t} \cdot \lambda_{k,i,j,n,t} \) \\ 
    Estimate and store fatigue prediction $F_{\text{pred,k,t}}$ using estimated parameters ($\mu_{k,t}$ or $\lambda_{k,i,j,t}$), last prediction $F_{\text{pred,k,t-1}}$, and fatigue model Eq. \ref{eq:fatigue};\\
    $\#\#\#\#\#$ {\bf Step two: estimated $\tau_{k,i,t}$ -- completion time of human $e^h_k$ to perform $task_{i}$; $F_{k,i, t+\tau_{k,i,t}}$ -- task-level fatigue prediction} \\
    \For{each human $e_{k}^h \in \mathcal{E}^h$} {
    $\#\#\#\#\#$ {Compute fatigue change for human $e_{k}^h$ when doing $task_i$} \\
        \For{each $task_i \in \mathcal{T}^h$} {
            \For{{each} $subtask_{i,j} \in task_i = \{subtask_{i,0}, subtask_{i,1}, \ldots, subtask_{i,j}, \ldots\}$} {
                Given outputs in step one:
                parameters $\mu_{k,t}$ and $\lambda_{k,i,j,t}$ for k-th human in doing $subtask_{i,j}$; \\
                Compute subtask completion time $subtime_{k,i,j,t}$ and subtask-level fatigue change $F_{k,t+subtime_{k,i,j,t}}$using Algorithm \ref{alg:subtime};\\
            }
            Compute task completion time $\tau_{k,i,t}$ and task-level fatigue change $F_{k,i,t+\tau_{k,i,t}}$ using Eq. \ref{eq:time} and Algorithm \ref{alg:subtime};\\
            {\bf Check fatigue violation given fatigue $F_{k,i,t+\tau_{k,i,t}}$;}
        }
    }
    {\bf Generate safe action set $A^{safe}$ and safe human set $E^{safe}$;}}
\end{algorithm*}

Within the second step of Algorithm~\ref{alg:PF_predict}, the $subtime_{k,i,j,t}$ of each subtask is computed using Algorithm \ref{alg:subtime}. This yields the prediction of the resulting fatigue increment, and the human fatigue state $F_{k,t + subtime_{k,i,j,t}}$. The overall ideal completion time of \(task_i\), \(\tau_{k,i,t}\), is then obtained as following:
\begin{equation} \label{eq:time}
\begin{gathered}
\tau_{k,i,t} = \sum \{subtime_{k,i,0,t}, \ldots, subtime_{k,i,j,t}, \ldots\}.
\end{gathered}
\end{equation}
The above equation omits several dynamic effects that are difficult to model deterministically, such as travel time between workstations and stochastic fluctuations in subtask execution, which were mentioned in Eq.\ref{eq:task_time_random}.
Nevertheless, it enable predicts task-level completion time \(\tau_{k,i,t}\) for each \(task_i \in \mathcal{T}^h\) and corresponding fatigue changes \(F_{k,i,t+\tau_{k,i,t}}\).
Furthermore, the task-level fatigue prediction enables the generation of a safe human set \( E^{safe} \) and action set \( A^{safe} \) that comply with fatigue constraints for PF-CD3Q decision-making. The safe human set is defined as:

\begin{equation} \label{eq:safe_human_set}
\begin{gathered}
    E^{safe} = \{E^{safe}_0, \ldots, E^{safe}_k, \ldots\}, \\
    E^{safe}_k = \{task_0, \ldots, task_i, \ldots\}, \\
    e^h_k \in \mathcal{E}^h, \, task_i \in \mathcal{T}^h,
\end{gathered}
\end{equation}
where \( E^{safe}_k \) comprises tasks the \( k \)-th human can undertake without violating fatigue constraints. Additionally, we generate a safe action set that includes task decisions feasible for at least one human entity:

\begin{equation} \label{eq:safe_action_set}
\begin{gathered}
    A^{safe} = \{task_0, \ldots, task_j, \ldots\}, \, task_j \in \mathcal{T}^h,
\end{gathered}
\end{equation}
where \( task_j \) denotes a verified, available task performable without breaching fatigue limits.

\subsection{Particle filter with CD3Q for task planning and allocation} \label{sec:PF_CD3Q_subsec}

As the PF-CD3Q outputs action \( a_t = \{task_i, e_k^h \text{ and/or } e_n^r\} \) (see Sec. \ref{sec:obj}), it follows a two-step process: first prioritizing the task to be executed, then allocating it to a human and/or robot. Both steps rely on task-level fatigue predictions. Initially, the safe RL strategy determines the first step, outputting \( a_t = \{task_i\} \). Subsequently, spatial-aware and fatigue-predictive task allocation expands the action, allocating suitable human and/or robot entities \(\{e_k^h \text{ and/or } e_n^r\} \) to complete the task.

\subsubsection{Constrained Markov decision process} \label{sec:CMDP}
We model the HRTPA process in production using a Constrained Markov Decision Process (CMDP) \cite{altman1993asymptotic}, which extends MDP by integrating constraints with rewards to optimize cumulative returns while keeping expected costs within specified limits \cite{altman2021constrained}. This is represented by the tuple \( G = \langle \mathcal{S}, \mathcal{A}, T, \mathcal{R}, \rho_0, \gamma, H, \mathcal{C}_0, \ldots, \mathcal{C}_k \rangle \), where \(\mathcal{S}\) is the state space, \(\mathcal{A}\) is the finite action space, \(\mathcal{R}\) is the reward function, \(T\) is the state transition function, \(\rho_0\) is the initial state distribution, \(\gamma \in (0, 1]\) is the discount factor, and \(H\) is the time horizon. The CMDP incorporates auxiliary cost functions \(\mathcal{C}_0, \ldots, \mathcal{C}_k\) for \( e^h_k \in \mathcal{E}^h \) (where \(\mathcal{E}^h\) is the human entity set, detailed in Sec. \ref{sec:entity}), where each \(C_k: \mathcal{S} \times \mathcal{A} \times \mathcal{S} \to \mathbb{R}\) maps transition tuples to the fatigue-related cost guiding the agent to avoid violating fatigue constraints.

At each time step \(t\), the agent observes state \(s_t \in \mathcal{S}\), receives reward \(r_t \in \mathcal{R}\) and costs \(c_{k,t} \in \mathcal{C}_k\), and selects action \(a_t \in \mathcal{A}\). The environment transitions to \(s_{t+1}\) via \(T(s_t, a_t, s_{t+1}) = P[s_{t+1}|s_t, a_t]\), where \(T: \mathcal{S} \times \mathcal{A} \times \mathcal{S} \to [0, 1]\) defines the transition probability. The agent then receives \(r_{t+1}\) and \(c_{k,t+1}\) for the new state.
The performance objective consists of the expected reward and cost returns: 

\begin{equation} \label{eq:CMDP_obj}
\begin{gathered}
    \max_{\pi_{\theta}} J(\pi_{\theta}), \\
    \text{s.t.} \quad \forall J_{C_k}(\pi_{\theta}) \leq d_k, \\
    J(\pi_{\theta}) = \mathbb{E}_\pi\left[\sum_{n=0}^{H} \gamma^n r_{t+n+1}\right], \\
    \quad J_{C_k}(\pi_{\theta}) = \mathbb{E}_\pi\left[\sum_{n=0}^{H} {\gamma}^n c_{k,t+n+1}\right],
\end{gathered}
\end{equation}
where \(d_0, \ldots, d_k\) are the constraint limits, and $\theta$ is the neural network of the policy. 

However, for the fatigue-constrained HRTPA problem in production, instead of designing cost functions and calculating the expected cost return \( J_{C_k}(\pi_\theta) \), we directly utilize \( F_{k,t} \) for \( e^h_k \in \mathcal{E}^h \), where \( F_{k,t} \) is the fatigue of the \( k \)-th human at time step \( t \) (see Sec. \ref{sec:entity}). The performance objective in Eq. \ref{eq:CMDP_obj} is thus transformed to:

\begin{equation} \label{eq:CMDP_obj_case}
\begin{gathered}
    \max_{\pi_\theta} J(\pi_\theta), \\
    \text{s.t.} \quad \forall e^h_k \in \mathcal{E}^h, \, F_{k,t} < d_k,
\end{gathered}
\end{equation}

For Eq. 10, the emphasis on cumulative expected costs necessitates a carefully designed cost function to guide agent learning, with its effectiveness critically influencing fatigue constraint performance. Conversely, Eq. \ref{eq:CMDP_obj_case} employs fatigue directly as a constraint, aligning more effectively with the objective in Sec. \ref{sec:obj} by ensuring human fatigue stays below threshold \( d_k \) at each time step.

\subsubsection{Objective function and loss} \label{sec:rl_obj_loss}

As outlined in Eq. \ref{eq:CMDP_obj_case}, the objective is to maximize the expected return while adhering to fatigue constraints. Following the Q-learning paradigm, which employs the off-policy Temporal Difference (TD) control \cite{watkins1989learning}, defined by:
\begin{align} \label{eq:TD_update}
q(s_t, a_t) & \leftarrow (1 - \alpha) q(s_t, a_t) \\
& + \alpha \left[ r_{t+1} + \gamma \argmax_{a'} q(s_{t+1}, a')) \right].
\end{align} 
This represents the TD update rule for the Q-value \( q(s_t, a_t) \) at time \( t \), combing the previous Q-value (weighted by \( 1 - \alpha \)) with \( \alpha \) times the TD target. The target includes the reward \( r_{t+1} \) plus a discounted future value, based on the maximum Q-value \( \argmax_{a'} q(s_{t+1}, a') \) over actions \( a' \) in state \( s_{t+1} \), scaled by \( \gamma \).
We then define the update rule for the constrained Q-value \( q^c_\theta(s_t, a_t) \), parameterized by the neural network \( \theta \):
\begin{equation} \label{eq:rl_update}
\begin{gathered} 
q^{c}_{\theta}(s_t, a_t) \leftarrow (1 - \alpha) q^{c}_{\theta}(s_t, a_t)
+ \alpha y_t,
\\ y_t =  r_{t+1} + \gamma q^{c}_{\theta^-}(s_{t+1}, \argmax_{a'\in A^{safe}(s_{t+1})} q^{c}_{\theta}(s_{t+1}, a')),
\end{gathered}
\end{equation}
where the TD target \( y_t \) incorporates the double Q-learning concept \cite{van2016deep}, reducing overestimation by decoupling action selection and evaluation in the max operation. It is computed as the reward \( r_{t+1} \) plus a discounted Q-value from the target network \( q^c_{\theta^-} \), evaluated at the next state \( s_{t+1} \) using the action \( a' \) that maximizes \( q^c_\theta(s_{t+1}, a') \) within the safe action set \( A^{safe}(s_{t+1}) \) (See Algorithm \ref{alg:PF_predict} and Eq. \ref{eq:safe_action_set} for details on the safe action set.). The target network, with parameters \( \theta^- \), replicates the online network, updating its parameters infrequently by copying from the online network to stabilize training and improve performance, and remains fixed otherwise.
Thus, the objective of the optimal policy, based on the Q-learning paradigm, is:
\begin{equation} \label{eq:rl_obj}
\begin{gathered}
    \pi^*_\theta(s) = \argmax_{a' \in A^{safe}(s)} q^{c*}_\theta(s, a').
\end{gathered}
\end{equation}
To obtain the optimal policy, we present the loss function \( L(\theta) \) and its gradient for training the Q-network:
\begin{equation} \label{eq:loss}
\begin{gathered}
L(\theta) = \mathbb{E}_{(s_t, a_t, s_{t+1}, r_{t+1}) \sim \mathcal{D}} \left[ \left( \alpha (y_t - q^c_\theta(s_t, a_t)) \right)^2 \right], \\
\nabla_\theta L(\theta) = \mathbb{E}_{(s_t, a_t, s_{t+1}, r_{t+1}) \sim \mathcal{D}} \left[ \alpha (y_t - q^c_\theta(s_t, a_t)) \nabla_\theta q^c_\theta(s_t, a_t) \right].
\end{gathered}
\end{equation}
The loss, formulated as a mean squared error (MSE), represents the expected squared difference between the target \( y_t \) and the predicted constrained Q-value \( q^c_\theta(s_t, a_t) \), averaged over the prioritized experience replay buffer \( \mathcal{D} \) \cite{schaul2015prioritized} and scaled by the learning rate \( \alpha \). This minimizes the discrepancy between the online and target networks by measuring the difference between the online network's predicted Q-values and the target Q-values. The gradient \( \nabla_\theta L(\theta) \), computed as the expected product of the TD error \( (y_t - q^c_\theta(s_t, a_t)) \) and the Q-value gradient with respect to \( \theta \), facilitates network optimization and updates the online network's parameters.

\begin{figure*}[htb] 
\centering	
	\includegraphics[width=0.99 \linewidth, height=0.48\linewidth]{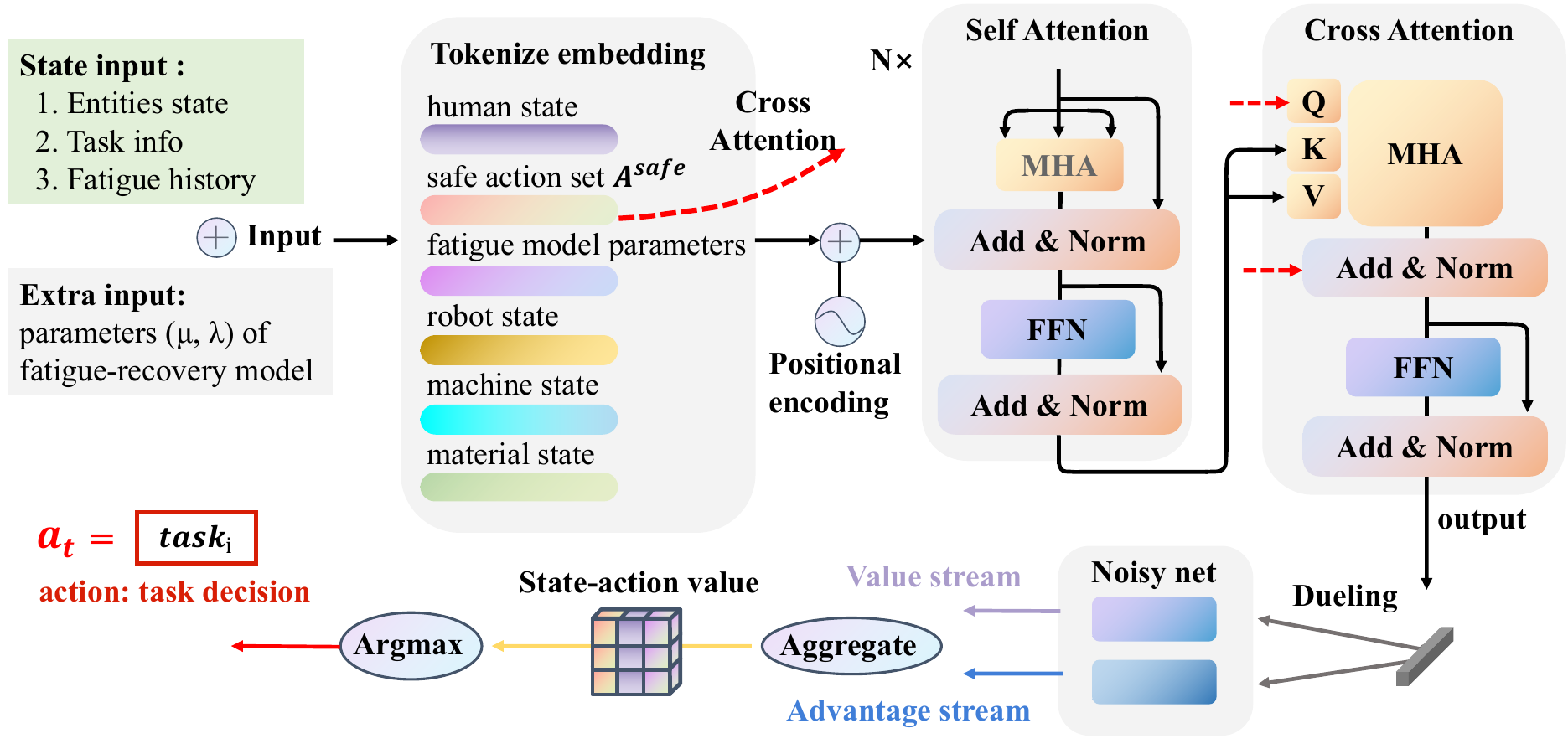}
    \caption{Network architecture of our method PF-CD3Q. \label{fig:network}} 
\end{figure*}

\subsubsection{Reward function} \label{sec:reward}
In Sec. \ref{sec:CMDP}, the objective transitions from Eq. \ref{eq:CMDP_obj} to Eq. \ref{eq:CMDP_obj_case}. Following Eq. \ref{eq:CMDP_obj_case}, we solve the constrained problem using task-level fatigue–prediction–based hard-action masking together with a single-objective reward function; no cumulative cost is optimized. We employ particle filters to predict task-level fatigue and define a safe action set \( A^{safe} \) to ensure the agent selects fatigue-constrained actions at each time step. This simplifies reward/cost function design, focusing solely on the makespan performance. Thus, the reward principle aims to keep the makespan within the maximum time horizon \( H \) while minimizing it, and to maximize the completion degree toward 1. Accordingly, we propose the following reward function:

\begin{equation} \label{equation:reward}
\begin{gathered}
    R = R_{time} + R_{goal} + R_{progress}, \\
    R_{time} = -\eta_1, \\
    R_{goal} = 
    \begin{cases}
        \eta_3, & \text{if new progress is made}, \\
        0, & \text{otherwise},
    \end{cases} \\
    R_{progress} = 
    \begin{cases}
        -\eta_2, & \text{if goal not done by end}, \\
        \eta_2, & \text{if goal done within time horizon}, \\
        0, & \text{otherwise},
    \end{cases}
\end{gathered}
\end{equation}
where \( \eta_1, \eta_2, \eta_3 \) are constant hyperparameters. The term \( r_{time} \) imposes a negative penalty as time elapses, encouraging a shorter makespan. A positive reward \( r_{goal} \) is granted if the production goal is achieved within or before the time horizon \( H \), while a negative reward applies if it is not met by the end. The \( r_{progress} \) term provides a positive reward for any progress toward the goal (e.g., product output), even if incomplete, to promote continuous advancement.

\subsubsection{Spatial and fatigue-predictive task allocation} \label{sec:expand_action}
As illustrated in Fig. \ref{fig:algorithm}, the PF-CD3Q produces action \( a_t = \{task_i, e_k^h \text{ and/or } e_n^r\} \) (see Sec. \ref{sec:obj}), following a two-step process. Starting with the task planning result \( a_t = \{task_i\} \) from the safe RL-based policy \( \pi_\theta \), we then select suitable human and/or robot candidates for \( task_i \). In the open workspace production environment, where movable entities navigate for various tasks, we implement a distance-greedy path-planning strategy, such as Dijkstra’s algorithm \cite{dijkstra2022note} or Hybrid A* \cite{montemerlo2008junior}. Leveraging task-level fatigue predictions, we filter eligible humans and choose the nearest entities, resulting in the final allocation output \( a_t = \{task_i, e_k^h \text{ and/or } e_n^r\} \) to the environment.

\subsection{Attention-based neural network design} \label{sec:neural_network}

The complete network architecture (\(\theta\)) of PF-CD3Q, depicted in Fig. \ref{fig:network}, comprises encoders for tokenized embeddings, the attention-based architecture for handling heterogeneous inputs, and the dueling network with Noisy Net for output projection. 

\subsubsection{State input embeddings}

To process heterogeneous state information, we treat each data component as a distinct token, encoding categorical inputs via embeddings and continuous values through linear layers. The process starts with real-time state data, including states of humans, robots, machines, materials, and production progress, structured as per Eq. \ref{eq:state}: \(\mathbf{s}_{t}' = \{\mathbf{s}^h_t, \mathbf{s}^r_t, \mathbf{s}^{mac}_t, \mathbf{s}^{mat}_t, \mathcal{T}, \mathcal{O}\} \subset \mathbf{s}_t\), where \(\mathcal{T}\) denotes the task set and \(\mathcal{O}\) the subtask set. We also encode fatigue model parameters (\(\lambda_{k,i,j}, \mu_k\) for \( e^h_k \in \mathcal{E}^h\), with \(\lambda_{k,i,j}\) corresponding to \( subtask_{i,j} \in task_i\)) for each human, capturing fatigue variations across subtasks. Each token is processed by multiple encoders defined as:

\begin{eqnarray}
    Embed(c_i) = \rho_i (c_i), \, c_i \in \mathbf{s}_t, \, \rho_i \subset \rho.
\end{eqnarray}

Here, each token \( c_i \in \mathbf{s}_t \) is handled by a dedicated encoder \(\rho_i\), part of the overall encoder network \(\rho\).

In practice, we utilize Multilayer Perceptrons (MLPs) or Embedding models as encoders due to their robust performance and low parameter complexity. Input features are mapped into a fixed-dimensional space, with positional encoding applied to preserve sequence information and differentiate heterogeneous data, computed as:

\begin{equation} \label{eq:posed_encode}
\begin{gathered}
PE(pos, 2i) = \sin\left(\frac{pos}{10000^{2i / d_{\text{model}}}}\right), \\
PE(pos, 2i+1) = \cos\left(\frac{pos}{10000^{2i / d_{\text{model}}}}\right), \\
c_{\text{posed}}(pos, i) = c_{\text{embedding}}(pos, i) + PE(pos, i),
\end{gathered}
\end{equation}
where \( pos \) denotes the token's position, \( i \) is the dimension index (with \( 2i \) and \( 2i+1 \) for even and odd indices), and \( d_{\text{model}} \) represents the model's dimensionality (e.g., embedding size). The encoded input is then formed accordingly.

\subsubsection{Attention-based feature processing}

Multi-Head Attention (MHA) layers process these varied encoded features \cite{vaswani2017attention}, optimized for handling diverse data. The MHA layers perform self-attention and cross-attention, defined as:

\begin{eqnarray}
    \operatorname{Attention}(Q, K, V) = \operatorname{softmax}\left(\frac{Q K^T}{\sqrt{d_k}}\right) V,
    \label{equation:architecture-attn}
\end{eqnarray}
where \(\operatorname{Attention}\) calculates a weighted sum of values (\(V\)) based on the dot product of queries (\(Q\)) and keys (\(K\)), scaled by \(\sqrt{d_k}\) for training stability, with \(d_k\) as the key vector dimension. We adopt the Transformer architecture, combining self-attention and cross-attention, reinforced by residual connections \cite{he2016deep}. Self-attention uses all encoded features as queries, keys, and values, enabling the model to detect intra-sequence relationships via attention weights. After multiple layers, the self-attention output feeds into the keys and values of the cross-attention layer. In cross-attention, the safe action set \(A^{safe}\) serves as the query, denoted as \(Q_{safe} = [q_{safe}] \in \mathbb{R}^{1 \times d_k}\), with keys (\(K\)) and values (\(V\)) from the self-attention output, yielding a fatigue-informed action output \(\phi\).

Referencing the dueling network \cite{wang2016dueling}, known to enhance Q-learning performance, the output \(\phi\) is split into a value stream and an advantage stream. This separation distinguishes state values from action advantages, addressing non-identifiability issues, and improving optimal action selection during policy evaluation. To enhance exploration during training, we incorporate Noisy Nets \cite{noisynet2017} into both streams, adding controlled randomness that surpasses traditional \(\epsilon\)-greedy methods. 
Finally, the q-network \( q^c_\theta(s) \), representing the action value for task decisions, generates the greedy action \( a_t = \argmax_{a'} q^c_\theta(s, a') \) (\( task_i \in \mathcal{T} \)). Subsequently, after the neural network outputs the task decision, \( a_t \) is expanded, as detailed in Sec. \ref{sec:expand_action}, using spatial-aware and fatigue-predictive task allocation.

\section{Experiments}

\begin{figure*}[ht] 
\centering	
	\includegraphics[width=0.98 \linewidth, height=1.2\linewidth]{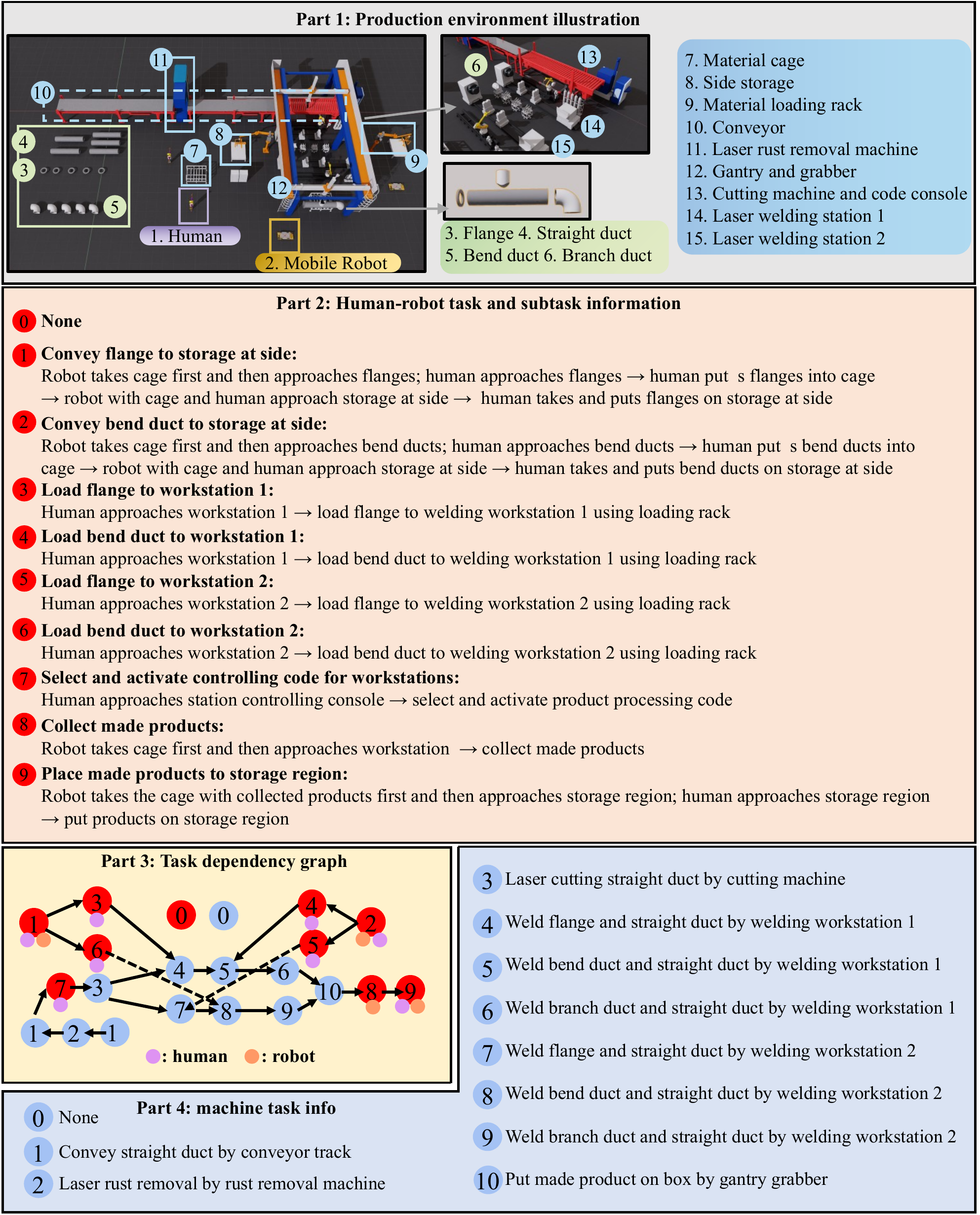}
\caption{Experimental environment, entities, human-robot-machine task descriptions, and task dependency graph \label{fig:exp_env}} 
\end{figure*}

In this section, we outline the problem scenario, evaluation metrics, and baseline algorithms for comparison. The objectives are dual:
(1) To determine if the proposed PF-CD3Q algorithm facilitates effective real-time HRTPA while maintaining human fatigue within safe limits, assessed by comparing it with vanilla reinforcement learning methods and representative safe RL baselines.
(2) To evaluate the prediction accuracy of fatigue states and model parameters using the PF-based estimator, in comparison with alternative methods such as the Kalman filter (KF).

\subsection{Experiment environment setup}

Fig. \ref{fig:exp_env} illustrates the detailed experimental setup for the HRTPA production problem, encompassing the simulation scenario, task and subtask definitions, and task dependency graph. The layout, 3D model, and task descriptions are derived from a real air conditioning duct production factory in modular construction.

The scenario features multiple entities, including humans, mobile robots, and stationary materials, machines, and tools. Specifically, producing a product involves four components: flange, straight duct, bending duct, and branch duct. To manufacture these, humans and robots must collaborate, utilizing machines and tools to complete tasks such as material preparation (transport and loading), material processing, and finished product collection and storage. The human-robot tasks and corresponding subtasks involve some light tasks handled by humans alone, while heavy workloads, particularly those involving the conveyance of heavy materials, require human-robot collaboration. Machine descriptions and capabilities are also presented.

Given the production context, the problem entails coordinating human-robot resources to maximize efficiency while adhering to safe human fatigue limits. The key challenge lies in determining when to perform which task and allocating the responsible agent (human and/or robot). Subtask sequencing exhibits minimal variability, eliminating the need for subtask decisions; once a task and its allocated human and/or robot are determined, subtasks are executed sequentially.

\subsection{Fatigue-related settings}
\begin{table}[htb]
\caption{Static and dynamic values of human fatigue model parameters, as detailed in Sec. \ref{sec:fatigue} and Sec. \ref{sec:method_pf_predict} for parameter definitions. \label{table:exp_fatigue_parameters} }
\centering
\begin{tabular}{p{5.3cm}c} 
\hline
    \begin{tabular}[p]{@{}l@{}} 1. (Static) $\lambda$--Fatigue of subtask\end{tabular}     & \begin{tabular}[p]{@{}l@{}}Value\end{tabular} \\
\hline
{Put flange into cage} & 0.12 \\
{Put bend duct into cage} & 0.18 \\
{Put flange on side storage} & 0.12 \\
{Put bend duct on side storage} & 0.18 \\
{Loading flange on welding station 1} & 0.36 \\
{Loading flange on welding station 2} & 0.36 \\
{Loading bend duct on welding station 1} & 0.45 \\
{Loading bend duct on welding station 2} & 0.45 \\
{Activate station controlling code} & 0.03 \\
{Place made product on storage} & 0.45 \\
\hline

    \begin{tabular}[p]{@{}l@{}}2. (Static) $\mu$--recovery state\end{tabular}     & \begin{tabular}[p]{@{}l@{}}Value\end{tabular} \\
\hline
    {Free} & 0.015 \\
    {Waiting} & 0.015 \\
    {Walking} & 0.006 \\
\hline

    \begin{tabular}[p]{@{}l@{}}3. $\delta_{eff}$--production efficiency change\end{tabular}     & \begin{tabular}[p]{@{}l@{}}Value=0.3\end{tabular} \\

\hline
    \begin{tabular}[p]{@{}l@{}}4. (Episodic) $\lambda$ for random human types\end{tabular}     & \begin{tabular}[p]{@{}l@{}}\end{tabular} \\
\hline
    \multicolumn{2}{l}{Human type: $\{\text{weak, normal, strong}\}$ = $\{\text{1.2, 1.0, 0.8}\}$ }\\
    \multicolumn{2}{l}{$\lambda$ = $\lambda$*(random human type)}\\
\hline
    \begin{tabular}[p]{@{}l@{}}5. Inaccurate initial fatigue model for algorithm \end{tabular}     & \begin{tabular}[p]{@{}l@{}}\end{tabular} \\
\hline
    \multicolumn{2}{l}{$\sigma_{init}$ = 0.2, see Eq. \ref{eq:fatigue_noise_param} for details}\\
\hline
    \begin{tabular}[p]{@{}l@{}}5. Fluctuated subtask completion time\end{tabular}     & \begin{tabular}[p]{@{}l@{}}\end{tabular} \\
\hline
    \multicolumn{2}{l}{$\sigma_{time}$ = 0.1, see Eq. \ref{eq:task_time_random} for details}\\
\hline
    \begin{tabular}[p]{@{}l@{}}6. Uniformly initialize particles within the parameter bounds \end{tabular}     & \begin{tabular}[p]{@{}l@{}}\end{tabular} \\
\hline
    \multicolumn{2}{l}{$\sigma_{particle}$ = 0.3, see Eq. \ref{eq:particle_initial} for details}\\
     \multicolumn{2}{l}{$N_p = 500$, the number of particles is 500} \\
\hline
    \begin{tabular}[p]{@{}l@{}}7. Fatigue measurement with Gaussian noise\end{tabular}     & \begin{tabular}[p]{@{}l@{}}\end{tabular} \\
\hline
    \multicolumn{2}{l}{$z_{k,t} \sim \mathcal{N}(F_{k,t}, \sigma_m^2), \sigma_m = 5\cdot10^{-5}$, where $F_{k,t}$ denotes the true }\\
    \multicolumn{2}{l}{fatigue value of the k-th human at time step t}\\
\hline
\end{tabular} 
\end{table}
Table \ref{table:exp_fatigue_parameters} details the parameter configuration of the fatigue-recovery model (see Sec. \ref{sec:fatigue} for further information). Drawing on prior research \cite{liu2023integration}, which accounts for varied fatigue change rates—slow, moderate, and fast—we establish static parameter values for different subtasks, tailored to their specific fatigue dynamics. At the start of each episode, we introduce variability by dynamically adjusting these parameters, randomizing \(\lambda\) across three human categories: weak, normal, and strong.
Per our Assumption \ref{ass:parameter}, which indicates that human fatigue sensitivity may fluctuate daily in real production, the algorithm initially lacks knowledge of the true \(\mu\) and \(\lambda\) 
values, relying instead on noisy estimates that necessitate real-time refinement. Consequently, as shown in Eq. \ref{eq:fatigue_noise_param}, we initialize \(\lambda\) and recovery \(\mu\) randomly within \(1 \pm 20\%\) of their true values, with $\sigma_{init} = 0.2$. During fatigue changes, production efficiency adjusts according to Eq. \ref{eq:efficiency}, with a fixed \(\delta_{eff}\) value of 0.3. Stochastic fluctuations in subtask completion time due to fatigue and other sources of uncertainty are modeled via Eq.\ref{eq:task_time_random}, with $\sigma_{time} = 0.1$. Particle initialization is specified in Eq.\ref{eq:particle_initial}, with $\sigma_{particle} = 0.3$. 

During the simulation of the production, the human operator’s true fatigue is generated using the fatigue–recovery model described in Eq. \ref{eq:fatigue} with ground-truth parameters. At each time step, the algorithm receives fatigue values with Gaussian measurement noise added: $z_{k,t} \sim \mathcal{N}(F_{k,t}, \sigma_m^2)$, where $\sigma_m = 5\cdot10^{-5}$.

\subsection{Training protocols and metrics}

The environment is established using NVIDIA's Isaac Sim 3D simulator \cite{liang2018gpu}. The algorithm is trained across numerous episodes, with random initialization at the start of each episode to capture production variability. This involves randomly allocating initial positions for human, robot, and cage entities, with the number of humans and robots varying randomly between 1 and 3.
The training and testing phases utilize different random seeds to ensure variability and assess the algorithm's adaptability. We employ the Adam optimizer \cite{kingma2015adam} to train all RL-integrated algorithms, with experiments conducted on a system featuring an Intel(R) Xeon(R) Platinum 8370C CPU and an NVIDIA GeForce RTX 4090.

For performance evaluation, we emphasize makespan, progress, and overwork as key metrics. Makespan quantifies the total time required to complete all tasks. Progress reflects the task completion level, ranging from 0 (no completion) to 1 (full completion of the manufacturing order) within the defined time horizon. To assess the algorithm's effectiveness in managing human fatigue, we calculate the average number of fatigue limit violation instances per production episode, referred to as overwork. For boxplots that include t-test statistics, a larger t-value and a smaller p-value indicate a more significant difference between the compared results.

\subsection{Comparison explanation}
\begin{table}[htb]
\caption{Comparison summary of RL or safe RL algorithms. Our proposed method is indicated by a ``$^\dagger$"\label{table:algo_comp}. }
\centering
\begin{tabular}{p{2cm}|c|c|c} 
\hline
    & \begin{tabular}[p]{@{}l@{}}Reward \\ penalty\end{tabular} & \begin{tabular}[c]{@{}l@{}} Cost \\ function \end{tabular} & \begin{tabular}[c]{@{}l@{}}Safe \\action set\end{tabular} \\
\hline
{DQN} &\checkmark & \scalebox{0.75}{\usym{2613}}   & \scalebox{0.75}{\usym{2613}} \\
{PPO} &\checkmark & \scalebox{0.75}{\usym{2613}}   & \scalebox{0.75}{\usym{2613}} \\
{D3QN} &\checkmark & \scalebox{0.75}{\usym{2613}}   & \scalebox{0.75}{\usym{2613}} \\
{PPO-Lag} & \scalebox{0.75}{\usym{2613}}& \checkmark & \scalebox{0.75}{\usym{2613}} \\
{PF-DQN}$^\dagger$ & \scalebox{0.75}{\usym{2613}} & \scalebox{0.75}{\usym{2613}} & \checkmark \\
{PF-PPO}$^\dagger$ & \scalebox{0.75}{\usym{2613}} & \scalebox{0.75}{\usym{2613}} & \checkmark \\
{PF-CD3Q}$^\dagger$ & \scalebox{0.75}{\usym{2613}} & \scalebox{0.75}{\usym{2613}} & \checkmark \\
{PF-PPO-Lag}$^\dagger$ & \scalebox{0.75}{\usym{2613}} & \checkmark  & \checkmark  \\

\hline
\end{tabular} 
\end{table}
The comparison is categorized into two main aspects: (1) makespan performance and the capability of RL or safe RL algorithms to prevent overwork, with a summary of the algorithms in Table \ref{table:algo_comp}. The details are:

DQN \cite{mnih2013playing}: A classic off-policy algorithm for discrete action spaces, suited to HRTPA's decision framework, with a fatigue-related reward penalty for overwork.

PPO \cite{schulman2017proximal}: A popular on-policy algorithm, stable in continuous spaces and adaptable to discrete ones, incorporating a fatigue-related reward penalty.

D3QN \cite{wang2016dueling}: An advanced Q-learning variant, enhanced by double Q-learning \cite{van2016deep} and dueling networks, with a fatigue-related reward penalty.

PPO-Lag \cite{ray2019benchmarking}: a safe RL algorithm, enhanced PPO using an additional cost function with a Lagrangian method for improved constraint guarantees.

PF-DQN, PF-PPO, PF-PPO-Lag, and PF-CD3Q: Extensions of DQN, PPO, PPO-Lag, and D3QN, respectively, integrating our particle filter-based fatigue prediction to create a safe action set, restricting exploration for fatigue compliance.

(2) The accuracy of the filter mechanism in estimating fatigue model parameters. For this comparison, we evaluate three filtering approaches: the Kalman filter (KF), the extended Kalman filter (EKF) \cite{welch1995introduction}, and the particle filter (PF) \cite{gustafsson2002particle}. The KF serves as an optimal estimator for linear systems, while the EKF extends this capability to nonlinear systems through local linearization. The PF uses a set of particles to approximate distributions, offering flexibility in modeling nonlinear and complex system dynamics.

\subsection{Performance in estimating fatigue model parameters}
\begin{figure*}[H]
\centering	
	\includegraphics[width=0.99 \linewidth, height=1.32\linewidth]{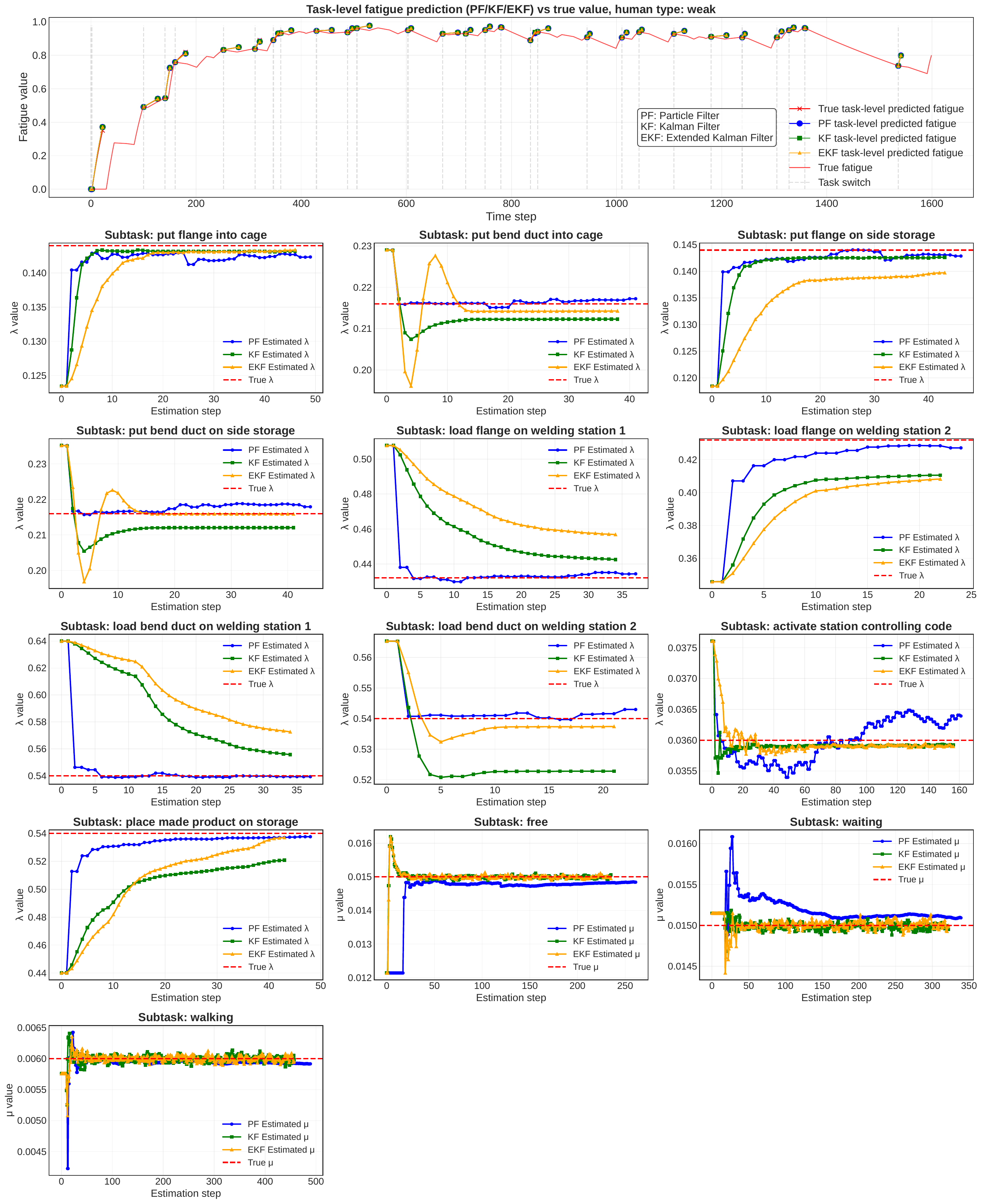}
\caption{Case study: Human episodic fatigue variation, fatigue model parameter estimation, and task-level fatigue prediction. \label{fig:filter_result}} 
\end{figure*}
\begin{figure*}[H]
\centering	
	\includegraphics[width=0.99 \linewidth, height=0.33\linewidth]{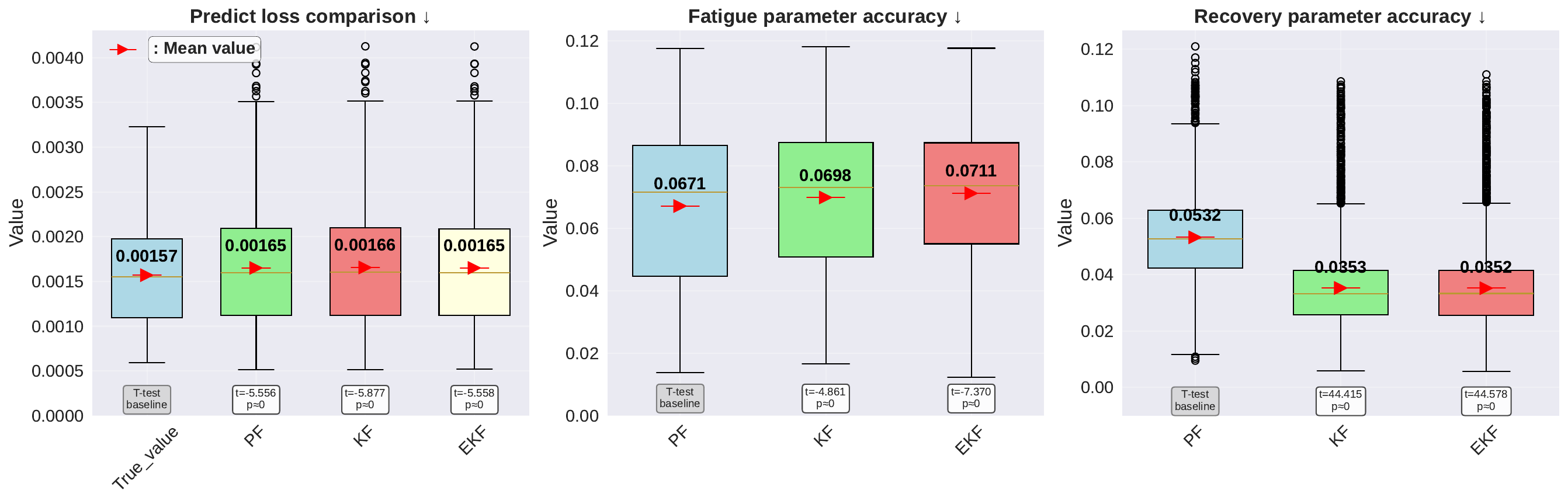}
\caption{Comparison of prediction loss and parameter accuracy for PF, KF, and EKF.\label{fig:filter_boxplot}} 
\end{figure*}

\begin{figure*}[htb]
\centering	
	\includegraphics[width=0.99 \linewidth, height=0.34\linewidth]{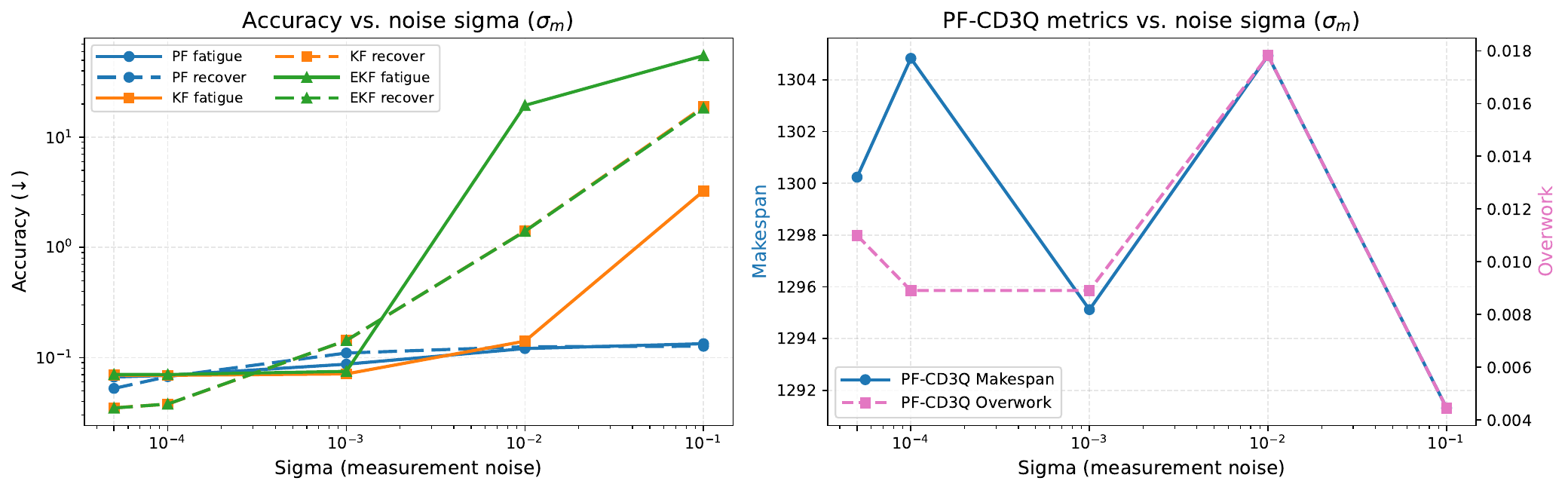}
\caption{Measurement noise analysis: filter accuracy and PF-CD3Q performance on test stage.\label{fig:filter_noise}} 
\end{figure*}

Fig. \ref{fig:filter_result} illustrates a case study on human episodic fatigue variation, fatigue model parameter estimation, and task-level fatigue prediction. The first subfigure displays the fatigue value change curve as a continuous red line, with vertical dashed lines marking human task switches. During production, fatigue typically rises and falls during recovery phases. At each task transition, the algorithm selects the next task and predicts the task-level fatigue change using the fatigue-recovery model and estimated parameters (detailed in Sec. \ref{sec:method_pf_predict}), estimating the fatigue increase if a human is allocated. For comparison, we display only the fatigue predictions for tasks subsequently executed by the human.
Both PF, KF, and EKF exhibit low prediction errors compared to using true fatigue model parameters (\(\lambda, \mu\)). However, even with accurate parameters, notable differences remain between predicted and actual production fatigue change curves. These discrepancies arise from: (1) task-level predictions assuming ideal completion times, overlooking human movement and waiting delays (which are challenging to account for due to the dynamic nature of production), and (2) the inherent random variability in the environment's task completion times. 

Fig. \ref{fig:filter_result} further displays the fatigue parameter estimation results. Each subtask is allocated a unique estimator to predict its specific fatigue parameters compared to true values. Note that updates to the estimation occur only when a human is performing the current subtask. Over time, as progress continues, the predictions become increasingly accurate.
The estimation is categorized into \(\lambda\) (fatigue accumulation) and \(\mu\) (recovery) predictions. For \(\lambda\) prediction, the particle filter (PF) generally exhibits faster convergence and higher accuracy. However, in the last three figures, KF and EKF outperform PF in estimating \(\mu\), particularly when the true value is close to zero, resulting in minimal fatigue changes across different parameter values. This leads to smaller differences in posterior probabilities among particles, reducing the impact of weight updates in PF.
Conversely, PF excels in handling complex nonlinear systems \cite{gustafsson2002particle}, as the fatigue model's accumulation formula is more intricate, while the recovery formula is relatively simpler. KF and EKF, being effective for linear or near-linear systems \cite{chen2003bayesian}, provide more accurate \(\mu\) predictions. However, our work prioritizes predicting fatigue increase and task-level fatigue changes, making the accuracy of \(\lambda\) estimation more critical.

Fig. \ref{fig:filter_boxplot} presents quantitative results on the accuracy of fatigue and recovery parameter estimation. The task-level predicted fatigue value loss is evaluated using the mean squared error (MSE) of the final fatigue difference post-task, ignoring variations between predicted and actual change curves.  Results show that filter-based predictions achieve low error relative to the ground-truth fatigue and recovery parameters, indicating that the deviation is acceptable.
Nonetheless, discrepancies persist even when using accurate parameters, primarily due to the idealized task-time model, which overlooks redundancy and execution uncertainty.
For fatigue parameters ($\lambda$) estimation, the PF achieves the lowest error (0.0671). The significance analysis (t-test) further confirms that PF performs significantly better than KF and EKF in estimating $\lambda$ (larger t-value and smaller p-value indicate greater difference). However, PF underperforms KF and EKF in recovery parameters ($\mu$) estimation.

Fig. \ref{fig:filter_noise} shows the measurement noise analysis, including the accuracy of three types of filters, PF, KF, and EKF, and PF-CD3Q performance during the test stage. The first subplot illustrates the estimation accuracy of the fatigue and recovery parameters using these three filters. When $\sigma_m < 10^{-4}$, all filters keep errors below 0.1. As noise increases, KF and EKF degrade and diverge, struggling to handle the higher noise levels, whereas PF maintains errors around 0.1. 
The second subplot presents the actual noise levels used in the experiments: $\sigma_m = 5 \times 10^{-5}$ during training, with deliberately higher values introduced during testing to assess robustness. The lower plots show PF-CD3Q performance in terms of makespan and cumulative overwork under these elevated noise conditions. Despite the significant increase in measurement noise, the PF-CD3Q exhibits only minor fluctuations in both metrics.

In summary, under the default low-noise setting ($\sigma_m = 5 \times 10^{-5}$), the PF outperforms both the KF and EKF in estimating the fatigue parameter $\lambda$, owing to its ability to model the fatigue dynamics' inherent nonlinearity. In contrast, KF and EKF achieve higher accuracy for the recovery parameter $\mu$, which follows a near-linear formula and typically takes small values near zero. This near-zero value poses a challenge for PF, as limited particle diversity results in minimal likelihood differences, hindering effective particle weighting and slightly reducing estimation accuracy for $\mu$. However, as measurement noise increases, KF and EKF rapidly degrade and eventually diverge, whereas PF maintains stable estimation accuracy. Consequently, the PF-CD3Q, which relies on PF-based task-level fatigue estimation, demonstrates robustness under high-noise conditions, exhibiting minor fluctuations in makespan and overwork.

\subsection{Performance of HRTPA Algorithms}
In this subsection, we evaluate whether our proposed PF-CD3Q algorithm supports fatigue-predictive HRTPA, optimizing makespan while preventing overwork, and compare its performance with other algorithms across various metrics.
We first present training-stage results, visualized through curves. Subsequently, we present quantitative test-stage outcomes through boxplots, bar charts, and tables. Additionally, we study the relationship between overwork/makespan and the number of humans and robots. Radar charts provide a visual comparison of algorithm performances. Finally, we use D3QN and PF-CD3Q as illustrative examples, presenting Gantt charts and fatigue change curves.

\subsubsection{Performance in training stage} \label{sec:train_stage}
\begin{figure*}[htp]
\centering	
	\includegraphics[width=0.99 \linewidth, height=0.62\linewidth]{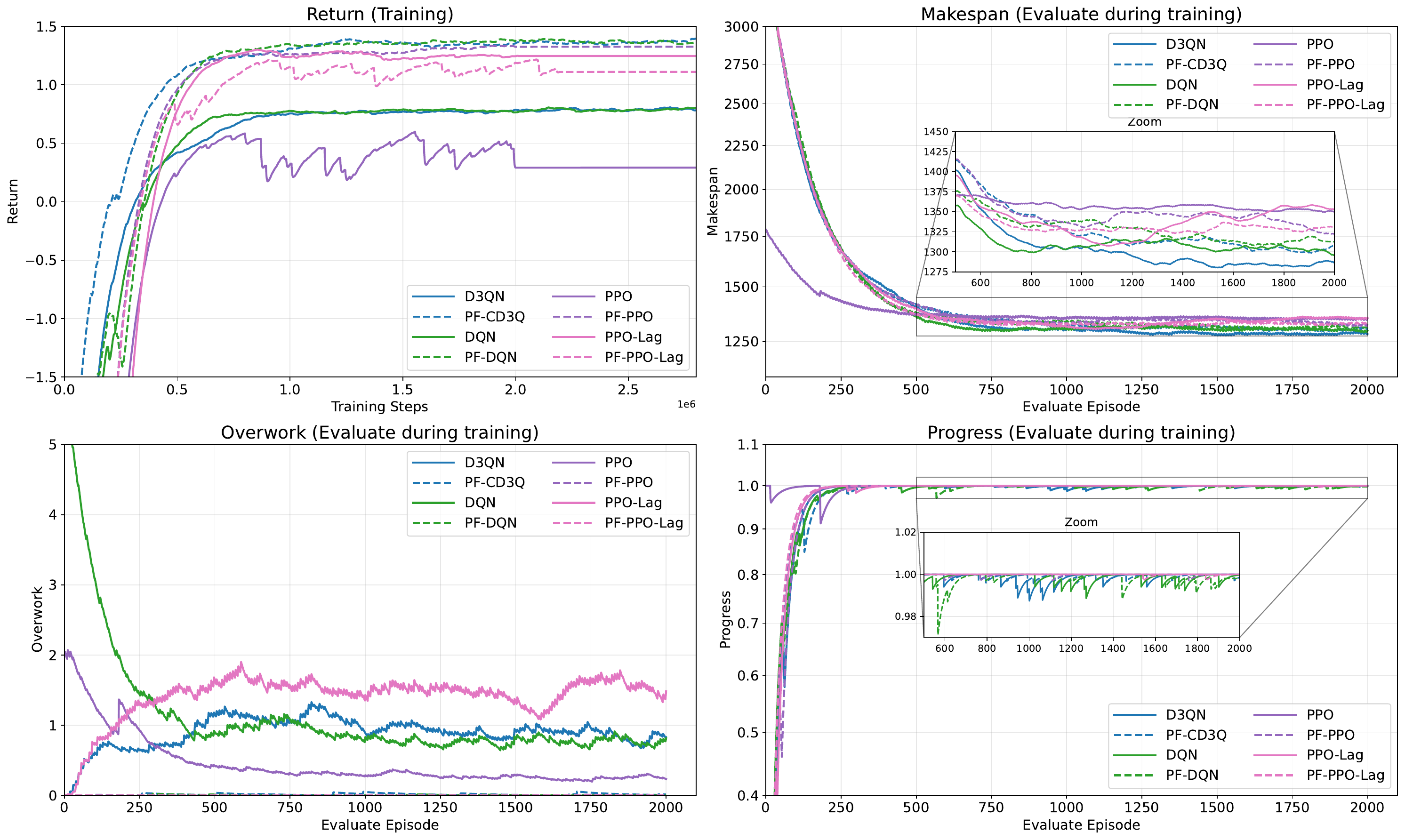}
\caption{Algorithms performance in training stage.\label{fig:training_curves}} 
\end{figure*}

Fig. \ref{fig:training_curves} illustrates training-stage performance across three key metrics. The first subplot shows reward, revealing a notable gap between PF-* and non-PF algorithms. As detailed in Table \ref{table:algo_comp}, non-PF algorithms (except PPO-Lag) employ overwork-related reward penalties during training, while PF-* algorithms avoid such adjustments, instead utilizing online filter-based task-level predictions to create safe action sets and enforce constrained decisions.
The second subplot displays makespan metrics, where D3QN achieves the best performance, with PF-CD3Q leading among PF-* algorithms. PPO exhibits the worst makespan curve, indicating poor efficiency.
The third subplot focuses on overwork metrics and shows that all PF-* algorithms maintain nearly zero overwork, indicating no fatigue violations during production. Among non-PF algorithms, PPO-Lag performs worst in overwork despite being a safe RL paradigm algorithm that uses a separate cost function with gradient-guided actor loss updates rather than reward penalties. PPO shows the lowest overwork among non-PF algorithms, but at the cost of poor makespan performance.
Regarding progress, all algorithms achieve nearly 100\% production completion rates, completing tasks within the given time horizon.

Key findings during training: (1) PF-* algorithms excel in overwork performance; (2) PPO and PPO-Lag perform poorly, with PPO-Lag surprisingly showing the worst overwork performance, likely due to ineffective cost function guidance in actor policy updates; (3) D3QN offers optimal makespan performance, while PF-CD3Q effectively balances strong makespan and overwork outcomes.

\subsubsection{Performance in test stage} \label{sec:test_stage}

\begin{figure*}[htb]
\centering	
	\includegraphics[width=0.99 \linewidth, height=0.42\linewidth]{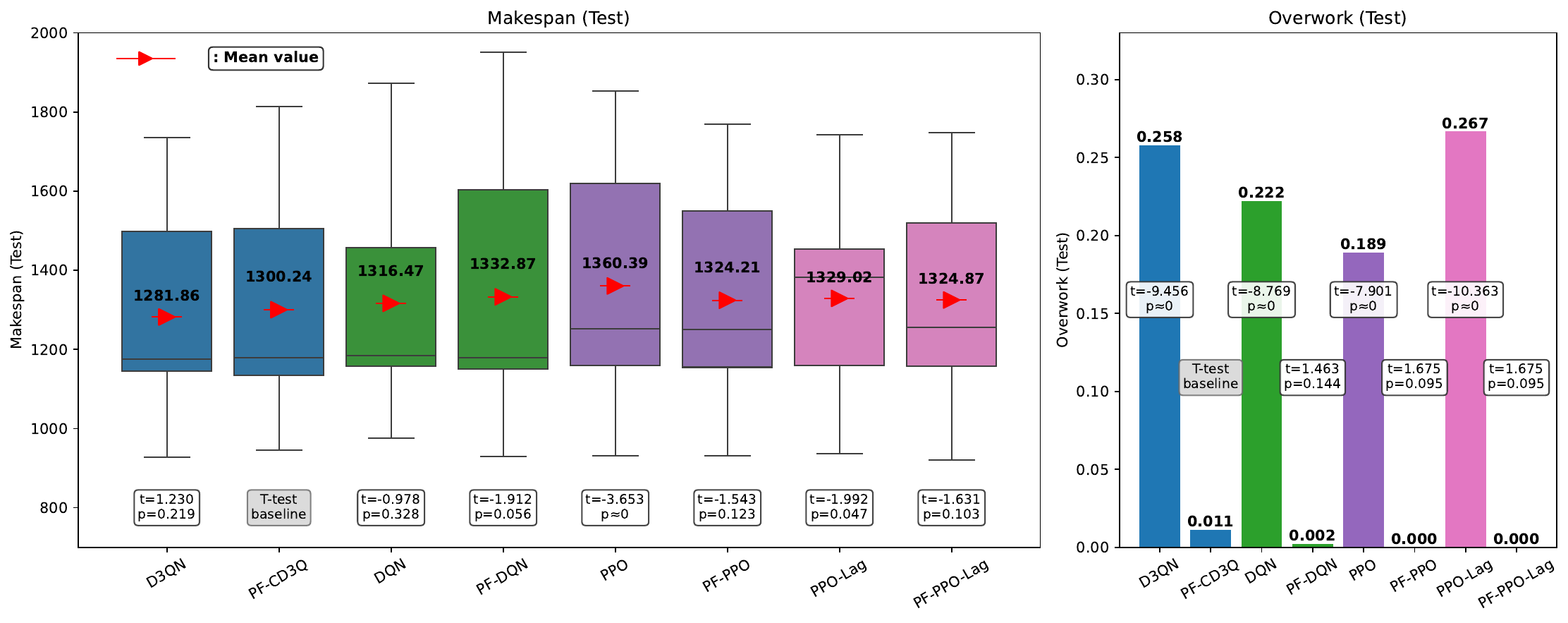}
\caption{Algorithm performance in the test stage, evaluated through makespan and overwork metrics.\label{fig:test_boxplot}} 
\end{figure*}
\begin{table*}[htbp]
\centering
\caption{Algorithm performance in the test stage, measured by makespan and overwork metrics. Note that \(^\dagger\) denotes the performance of our proposed algorithm. 'Hn' indicates n humans, and 'Rn' signifies n robots during testing.}\label{tab:combined_test}
\begin{tabular}{ccccccccccc}
\hline
Algorithm & H1,R1 & H1,R2 & H1,R3 & H2,R1 & H2,R2 & H2,R3 & H3,R1 & H3,R2 & H3,R3 & Mean \\
\hline
\multicolumn{11}{c}{\textbf{Makespan}} \\
\hline
DQN & 1764.60 & 1490.50 & 1587.98 & 1240.20 & 1172.34 & \textbf{1163.80} & 1180.70 & 1121.18 & 1126.90 & 1316.47 \\
PPO & 1615.76 & 1734.76 & 1738.04 & 1217.74 & 1215.42 & 1286.34 & 1149.90 & 1143.76 & 1141.80 & 1360.39 \\
D3QN & 1635.94 & 1506.26 & 1511.24 & 1193.52 & \textbf{1168.44} & 1174.56 & 1084.46 & 1129.10 & 1133.18 & \textbf{1281.86} \\
PPO-Lag & 1656.24 & \textbf{1451.18} & \textbf{1469.60} & 1401.06 & 1174.70 & 1170.62 & 1393.86 & 1118.38 & 1125.54 & 1329.02 \\
PF-DQN & 1767.10 & 1644.94 & 1641.90 & \textbf{1161.74} & 1173.22 & 1194.26 & 1174.22 & \textbf{1113.86} & \textbf{1124.62} & 1332.87 \\
PF-PPO & \textbf{1594.96} & 1603.56 & 1603.56 & 1215.74 & 1223.78 & 1275.72 & 1128.72 & 1128.28 & 1143.56 & 1324.21 \\
PF-CD3Q & 1638.96 & 1505.94 & 1549.68 & 1293.28 & 1192.80 & 1172.80 & \textbf{1084.22} & 1130.50 & 1133.98 & 1300.24 \\
PF-PPO-Lag & 1598.56 & 1589.68 & 1593.90 & 1222.22 & 1217.68 & 1286.16 & 1127.72 & 1136.86 & 1151.02 & 1324.87 \\
\hline
\multicolumn{11}{c}{\textbf{Overwork}} \\
\hline
DQN & 0.56 & 0.68 & 0.62 & 0.14 & 0.00 & 0.00 & 0.00 & 0.00 & 0.00 & 0.222 \\
PPO & 0.46 & 0.52 & 0.56 & 0.16 & 0.00 & 0.00 & 0.00 & 0.00 & 0.00 & 0.189 \\
D3QN & 0.56 & 0.86 & 0.84 & 0.00 & 0.04 & 0.02 & 0.00 & 0.00 & 0.00 & 0.258 \\
PPO-Lag & 0.56 & 1.00 & 0.84 & 0.00 & 0.00 & 0.00 & 0.00 & 0.00 & 0.00 & 0.267 \\
PF-DQN & 0.00 & 0.00 & 0.00 & 0.02 & 0.00 & 0.00 & 0.00 & 0.00 & 0.00 & 0.002 \\
PF-PPO & 0.00 & 0.00 & 0.00 & 0.00 & 0.00 & 0.00 & 0.00 & 0.00 & 0.00 & \textbf{0.000} \\
PF-CD3Q$^\dagger$ & 0.00 & 0.00 & 0.00 & 0.10 & 0.00 & 0.00 & 0.00 & 0.00 & 0.00 & 0.011 \\
PF-PPO-Lag & 0.00 & 0.00 & 0.00 & 0.00 & 0.00 & 0.00 & 0.00 & 0.00 & 0.00 & \textbf{0.000} \\
\hline
\end{tabular}
\end{table*}

\begin{figure*}[htb]
\centering	
	\includegraphics[width=0.99 \linewidth, height=0.62\linewidth]{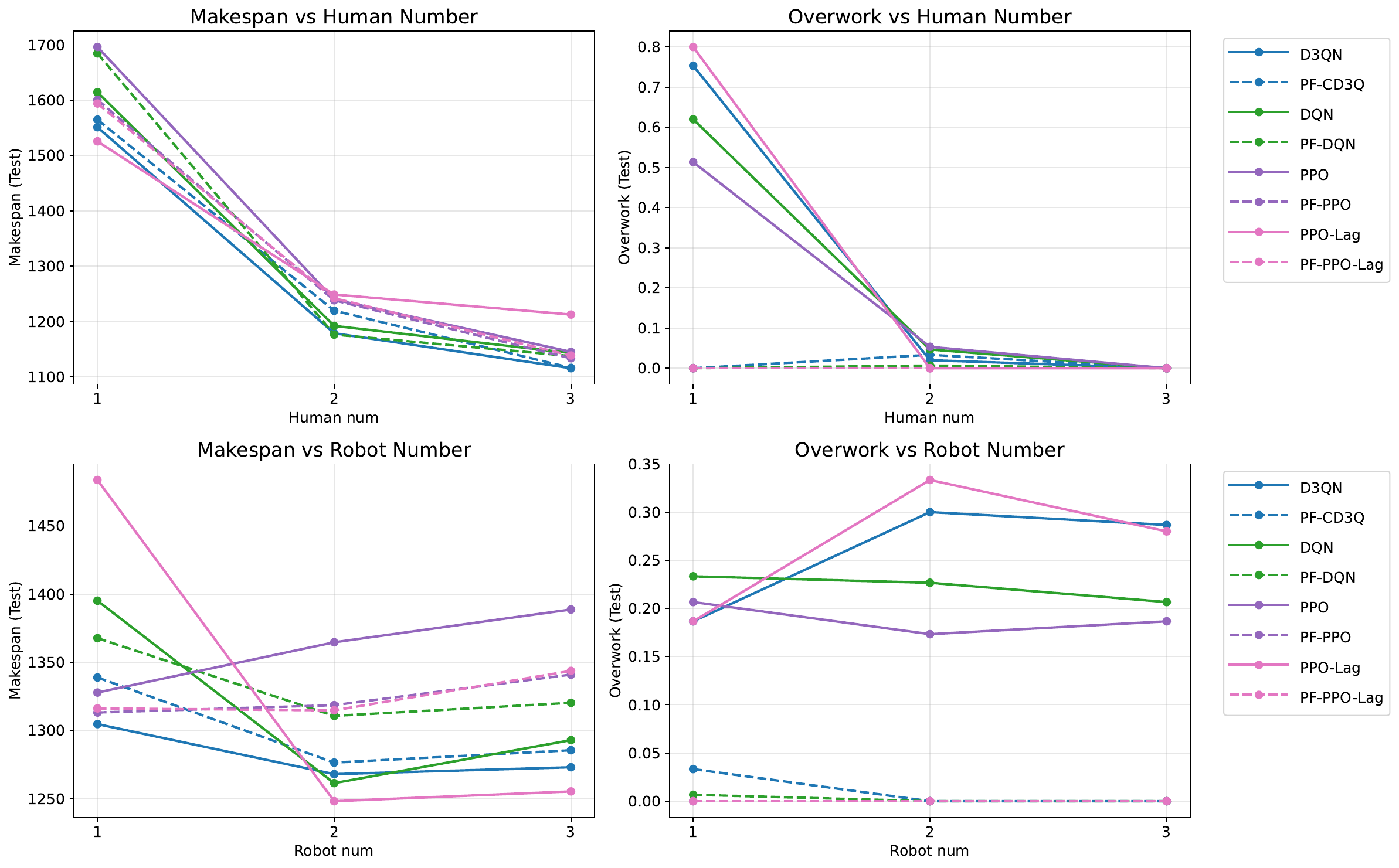}
\caption{Algorithm performance in various human-robot settings.\label{fig:test_human_robot_curves}} 
\end{figure*}

\begin{figure*}[htb]
\centering	
	\includegraphics[width=0.95 \linewidth, height=0.6\linewidth]{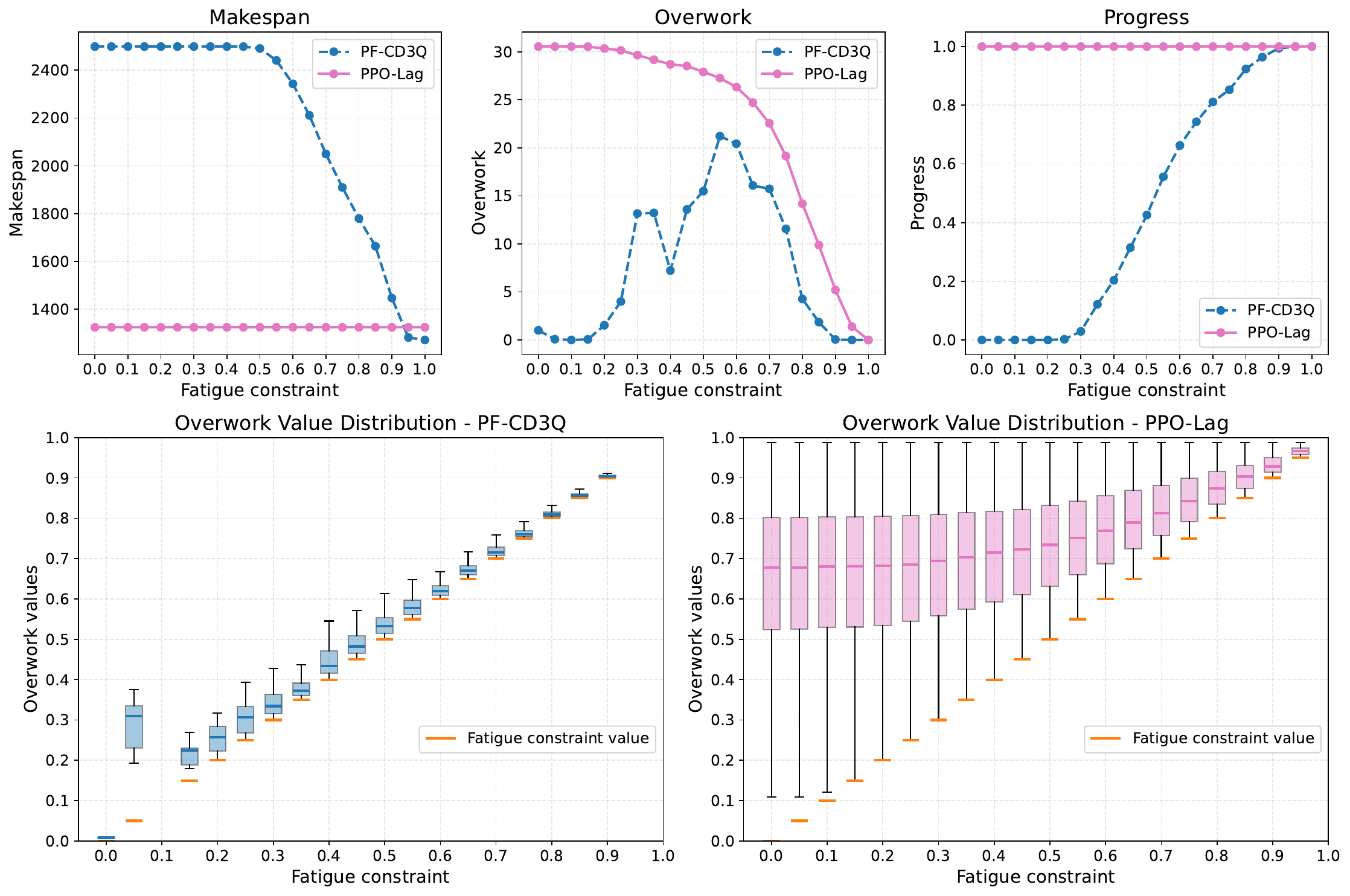}
\caption{Sensitivity analysis of fatigue constraints: comparison of PF-CD3Q and PPO-Lag. \label{fig:fatigue_sensitivity}} 
\end{figure*}

\begin{figure}[htb]
\centering	
	\includegraphics[width=0.9 \linewidth, height=0.72\linewidth]{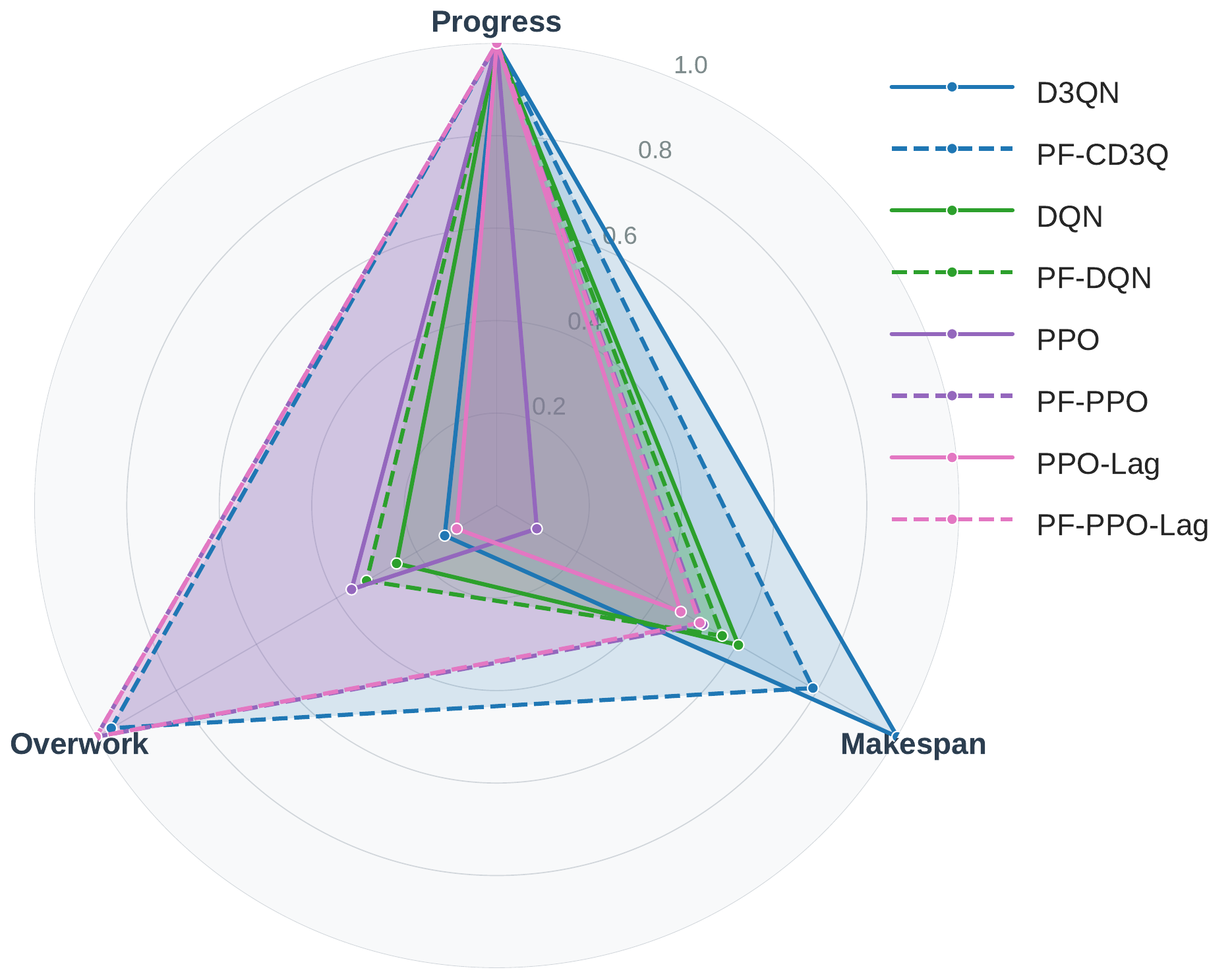}
\caption{Multi-metric algorithm ability: radar chart visualization.\label{fig:radar_chart_comprehensive}} 
\end{figure}
We provide more comprehensive and quantitative results for the test stage. Fig. \ref{fig:test_boxplot} visualizes the makespan and overwork performance. For brevity, we omit the progress metric as all algorithms achieve 100\% production task completion. The test-stage performance aligns with training-stage observations.
Makespan performance: D3QN performs best with 1281.86, followed by PF-CD3Q with 1300.24. PPO performs the worst with 1360.39. For Q-learning paradigms, incorporating particle filters worsens makespan performance; however, for PPO-based paradigms, the situation is reversed—incorporating PF shows no degradation in makespan. This occurs because PPO and PPO-Lag learn less effective policies, making the PF integration's impact on the makespan less pronounced. The t-test results further support these observations. Although D3QN outperforms PF-CD3Q in makespan, its p-values (0.219) indicate that the differences are not statistically significant. PF-CD3Q shows a noticeable makespan gap compared with most of the other algorithms (except DQN and D3QN). Overall, although incorporating PF may slightly reduce makespan performance, the degradation is modest, and PF-CD3Q remains close to the best-performing algorithm (D3QN).
For overwork performance: In contrast, D3QN exhibits the highest overwork rate, while PPO shows the lowest overwork among non-PF algorithms. Incorporating PF significantly reduces fatigue violation occurrences across all algorithms.

Table \ref{tab:combined_test} and Fig. \ref{fig:test_human_robot_curves} provide more detailed statistical results across different human-robot combinations. We observe that increasing the number of humans generally reduces both makespan and overwork across all algorithms. However, the improvement shows diminishing returns: transitioning from 1 to 2 humans yields significant improvements in both metrics, while increasing from 2 to 3 humans shows minimal additional benefits.
Increasing the number of robots shows mixed effects: it slightly decreases makespan for most algorithms, except for PPO, PPO-Lag, and PF-PPO-Lag, which show limited improvement.

Notable performance patterns: When considering configurations with 1 human and 1-3 robots, PF-CD3Q performs best among PF-* algorithms with an average makespan of 1564.86, followed by PF-DQN (1684.65), PF-PPO (1600.69), and PF-PPO-Lag (1594.04). PF-PPO achieves the best makespan performance in the 1-human, 1-robot configuration while maintaining zero overwork. PPO-Lag shows optimal makespan performance in configurations with 1 human and 2-3 robots, but at the cost of the highest overwork (1.0 and 0.84).
Considering average performance across all human-robot combinations, D3QN performs best in makespan, followed by PF-CD3Q.

Key Findings: (1) Test results are consistent with training-stage observations; (2) In this experimental setup, increasing the number of humans from 1 to 2 substantially lowers makespan and overwork, though additional increases yield diminishing returns; (3) While various algorithms perform best with specific human-robot combinations, D3QN delivers the highest average makespan performance, with PF-CD3Q placing second in makespan while achieving nearly optimal overwork results. Among PF-* algorithms in 1-human configurations, PF-CD3Q exhibits the best average performance, maintaining a zero overwork value.

\subsubsection{Sensitivity analysis of PF-CD3Q across fatigue thresholds} \label{sec:sensitivity}
We added a comprehensive sensitivity experiment that varies the fatigue constraint from 0.0 to 1.0 in increments of 0.05. For each setting, we test configurations with 1–3 human/robot entities, run ten randomized trials, and summarize the results using the completion time, overwork, and progress metrics. The finite completion time range is 2500. The new results are presented in Fig. \ref{fig:fatigue_sensitivity}.
From the results, when $d_k$ > 0.9, PF-CD3Q can complete production orders within the time frame without violating fatigue limits. Importantly, across the full range of fatigue constraints, PF-CD3Q consistently yields lower overwork than PPO-Lag. Although both algorithms are trained under the same constraint value (0.95), PF-CD3Q adapts well under different fatigue limits, effectively reducing overwork.
In contrast, PPO-Lag shows limited adaptability, producing unchanged makespan values across all settings and failing to reduce overwork as the fatigue constraint changes.
When the fatigue constraint becomes tight ($d_k$ < 0.9), PF-CD3Q may exhibit slight constraint violations; however, the overwork values remain close to the threshold. PPO-Lag, in comparison, frequently exhibits overwork near 1.0 across constraint levels, indicating weak constraint adherence.
In summary, the sensitivity analysis demonstrates that our approach remains robust across a wide range of fatigue-constraint settings, with PF-CD3Q showing strong adaptability and consistently lower overwork.

\subsubsection{Comprehensive analysis of experiment results} \label{sec:analysis_test}
As Sections \ref{sec:train_stage} and \ref{sec:test_stage} provide comprehensive results and key findings, as illustrated in Fig. \ref{fig:radar_chart_comprehensive}, this section analyzes these key findings and presents our conclusions.
(1) Overwork performance improvement of PF-*: We observe notable improvements in overwork performance after incorporating online filter-based task-level fatigue prediction to generate safe action sets. This improvement occurs because the method excludes unsafe behaviors that might lead to fatigue violations at each decision step. In contrast, the safe RL algorithm, PPO-Lag, employs soft constraints, indirectly restricting behavior through cost function gradients during actor function updates, and shows unsuccessful performance in fatigue-predictive scenarios. This might be due to the difficulty in cost function design, where cost violation occurrences can be sparse, potentially leading to unstable training and loss updates. 
(2) PPO vs. D3QN performance: PPO underperforms compared to D3QN in our context, as it is primarily designed for high-dimensional and continuous action spaces. Although PPO can manage discrete action spaces, it demands more careful tuning and strategy development than D3QN \cite{schulman2017proximal, de2024comparative}. In contrast, D3QN is specifically designed for discrete action space problems, and the omission of extra actor networks makes it more efficient for training and stability, making it suitable for our HRTPA problem domain.
(3) Optimal human workforce number: For different production problems, there exists an optimal number of human workers. In this study, increasing the number of workers from one to two significantly reduces the makespan and overwork. However, further increases yield diminishing returns, with minimal additional benefits.
(4) Comprehensive multi-metric performance: As shown in Fig. \ref{fig:radar_chart_comprehensive}, considering comprehensive multi-metric ability, PF-CD3Q emerges as the best performer. It achieves the second-best makespan while maintaining nearly zero overwork. Notably, in configurations with 1 human and 1-3 robots, PF-CD3Q maintains zero overwork and the lowest makespan among all PF-* algorithms.

\subsubsection{Case study of HRTPA algorithms} \label{sec:analysis_test}
\begin{figure*}[htb]
\centering	
	\includegraphics[width=0.99 \linewidth, height=0.98\linewidth]{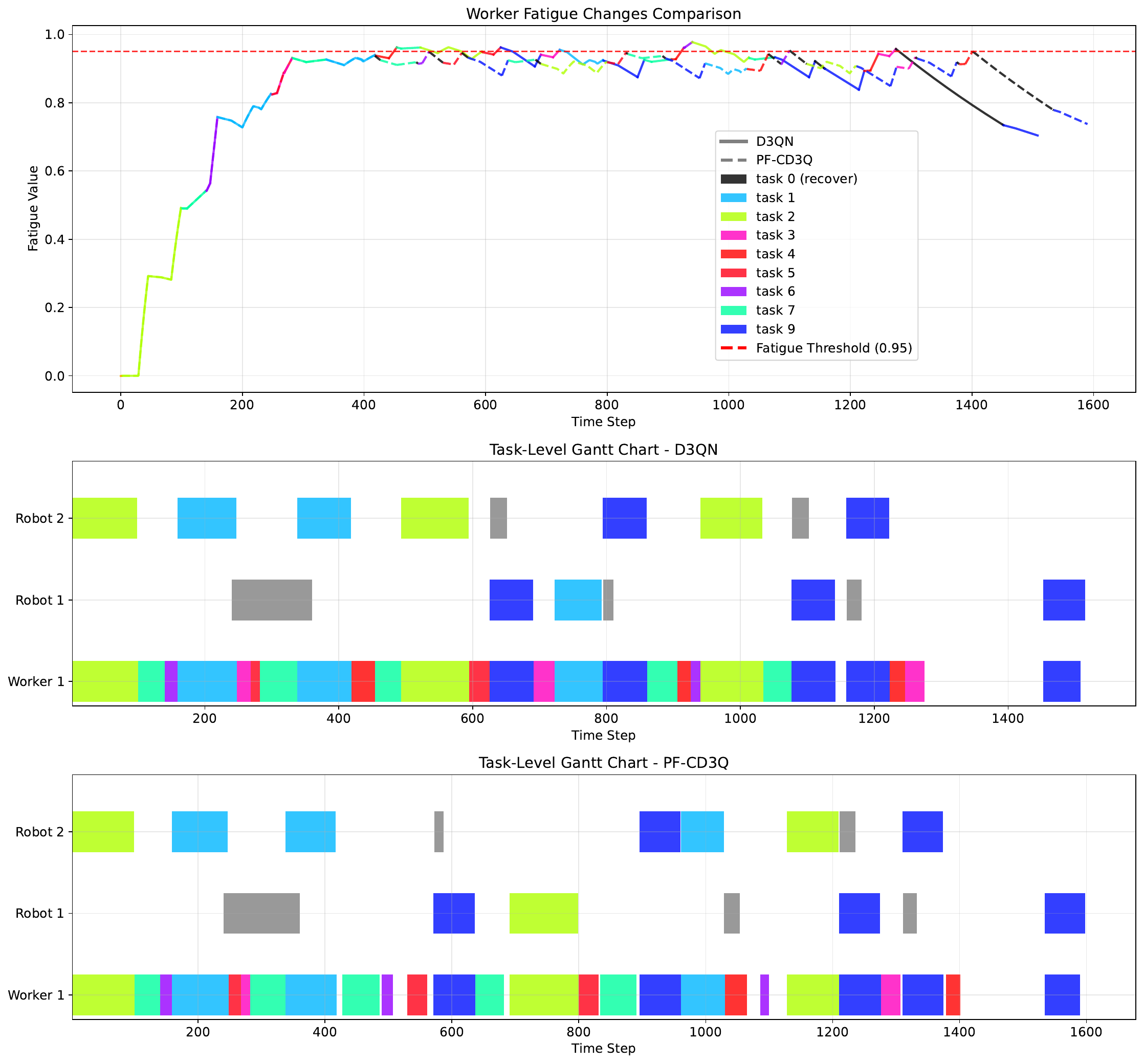}
\caption{PF-CD3Q vs D3QN: Case study of real-time HRTPA in the production process.\label{fig:gantt_chart_comparison}} 
\end{figure*}
As shown in Fig. \ref{fig:gantt_chart_comparison}, we finally showcase the real-time results of an episode HRTPA scenario with fatigue value changes, where the number of humans is 1, and robots are set to 2. For the fatigue curves, we use different colors to represent different tasks, and these colors are consistent with the color scheme in the Gantt chart: PF-CD3Q (fatigue curve shown as dashed lines) vs D3QN (fatigue curve shown as solid lines).
We observe that during the initial stage of the production process, the fatigue curves of these two algorithms show similar trends. However, when approaching the fatigue threshold, their decision-making strategies diverge significantly. For example, after 400 time steps, D3QN decides to perform task 4 immediately after task 1 with minimal idle free time. In contrast, PF-CD3Q allocates more rest time for the human worker and chooses to perform task 7 instead.
Subsequently, we observe that for PF-CD3Q, the human worker has more idle free time for recovery compared to D3QN, but this results in a longer makespan. In contrast, D3QN shows several instances of fatigue violations for the human worker.
In summary, this case study demonstrates how our PF-CD3Q algorithm facilitates real-time HRTPA in dynamic production scenarios while maintaining effective fatigue management.

\section{Discussion}

This work presents the particle filter with constrained dueling double deep Q-learning (PF-CD3Q) algorithm as a novel solution to the HRTPA problem. The experimental results highlight several important advantages over existing methods. First, the PF-based online filter \cite{gustafsson2002particle} enables real-time parameter updates with low latency and demonstrates strong robustness to measurement noise, making it suitable for dynamic production environments. Second, by integrating online filtering with safe RL, the proposed approach effectively handles scenarios where fatigue-related hyperparameters are initially inaccurate, whereas most existing work assumes full knowledge of the fatigue–recovery model parameters \cite{cai2023task}. Third, PF-CD3Q incorporates explicit fatigue constraints through effective fatigue-predictive decision-making, outperforming algorithms such as PPO-Lag \cite{ray2019benchmarking}, and can adapt to varying fatigue-limit settings in the test stage. Fourth, the ablation study confirms that the architectural components of our network design, including dueling networks \cite{wang2016dueling}, noisy layers \cite{noisynet2017}, and cross-attention \cite{vaswani2017attention}, collectively, contribute to improved decision-making performance, validating the architecture’s effectiveness for real-time HRTPA.

Despite these strengths, this study also has limitations. The proposed approach is evaluated within a limited type of production scenario, and validation in real-world human–robot collaborative systems has not yet been conducted. Deploying PF-CD3Q in physical environments will require accurate real-time fatigue monitoring and a robust digital-twin infrastructure, both of which require further research and development. These considerations point to important future directions for expanding the applicability and robustness of the algorithm in practical settings.

\section{Conclusion}

This work addresses the human-robot task planning and allocation (HRTPA) problem in dynamic production scenarios, tackling key challenges including real-time decision, physical fatigue constraints, varying worker efficiency, and inaccurate fatigue model hyperparameters. We present the \textbf{p}article \textbf{f}ilter with \textbf{c}onstrained \textbf{d}ueling \textbf{d}ouble \textbf{d}eep \textbf{Q}-learning (PF-CD3Q), a real-time fatigue-predictive HRTPA algorithm that integrates explicit fatigue constraints through a safe RL paradigm. Our key contributions include: (1) the first application of safe RL to HRTPA, ensuring fatigue-predictive and dynamic HRTPA, and can adapt to unseen fatigue constraints after training; (2) the particle filter (PF)-based online fatigue estimators that enable real-time fatigue tracking and generate safe action sets, and exhibit robustness to measurement noise; and (3) an attention-based Transformer architecture designed for processing heterogeneous data, including fatigue-related data, enhancing ergonomics-informed decision-making. Extensive experiments demonstrate that PF-based estimators achieve high accuracy in fatigue parameter estimation and task-level fatigue prediction. The PF-CD3Q algorithm excels in the overwork metric by utilizing online filter-based predictions to generate safe action sets, while securing the second-best makespan, resulting in the best overall performance across multiple metrics. In addition, the test-stage sensitivity analysis shows that PF-CD3Q generalizes well to unseen fatigue-constraint settings.

Limitations include the need to examine broader production environments, and real-world deployment has not yet been validated. Applying our framework to physical human–robot systems will require reliable real-time fatigue monitoring and a robust digital-twin infrastructure, both of which must be further developed.
Future research directions include: (1) expanding algorithm applicability to diverse manufacturing scenarios, including assembly lines, warehouse operations, and other collaborative manufacturing environments, to enhance the algorithm's adaptability and robustness across different production contexts; (2) addressing real-world implementation challenges, particularly the integration of wearable devices and perception technologies for real-time fatigue monitoring, including the development of reliable sensor fusion techniques and robust fatigue estimation models; (3) improving task description efficiency and algorithm scalability through the development of standardized task representation frameworks and automated task decomposition methods that can handle complex, multi-step manufacturing processes; (4) emergencies in production environments, including the implementation of online adaptation mechanisms, uncertainty quantification methods, and fallback strategies that can maintain system safety and performance under unexpected conditions.

\section*{CRediT authorship contribution statement}
\textbf{Jintao Xue}: Methodology, Software, Data curation, Writing – original draft. \textbf{Xiao Li}: Conceptualization, Supervision, Writing – review $\&$ editing. \textbf{Nianmin Zhang}: methodology analysis, experiment design.

\section*{Declaration of Competing Interest} The authors declare that they have no known competing financial interests or personal relationships that could have appeared to influence the work reported in this paper. 

\section*{Data availability} Data will be made available on request.

\section*{Acknowledgments} 
The work described in this paper is supported by grants from Technology Cooperation Funding Scheme (TCFS) (Ref No.GHP/321/22SZ), The University of Hong Kong (Ref No.109002002), and Innovation and Technology Fund (ITF) (Ref No. TP/041/24LP).

\bibliographystyle{cas-model2-names}

\bibliography{cas-refs}

\clearpage
\appendix

\section{Implementation Details and Additional Results}
\label{app:details}

Detailed hyperparameters of the training process are summarized in Table \ref{tab:appendix_train}. 
The sum-up task-level fatigue prediction latency of the proposed particle filter (PF) is approximately 20 $\mu$s, while the neural-network inference time is around 2 ms. 

\begin{figure*}[b]
\centering	
	\includegraphics[width=0.97 \linewidth, height=0.36\linewidth]{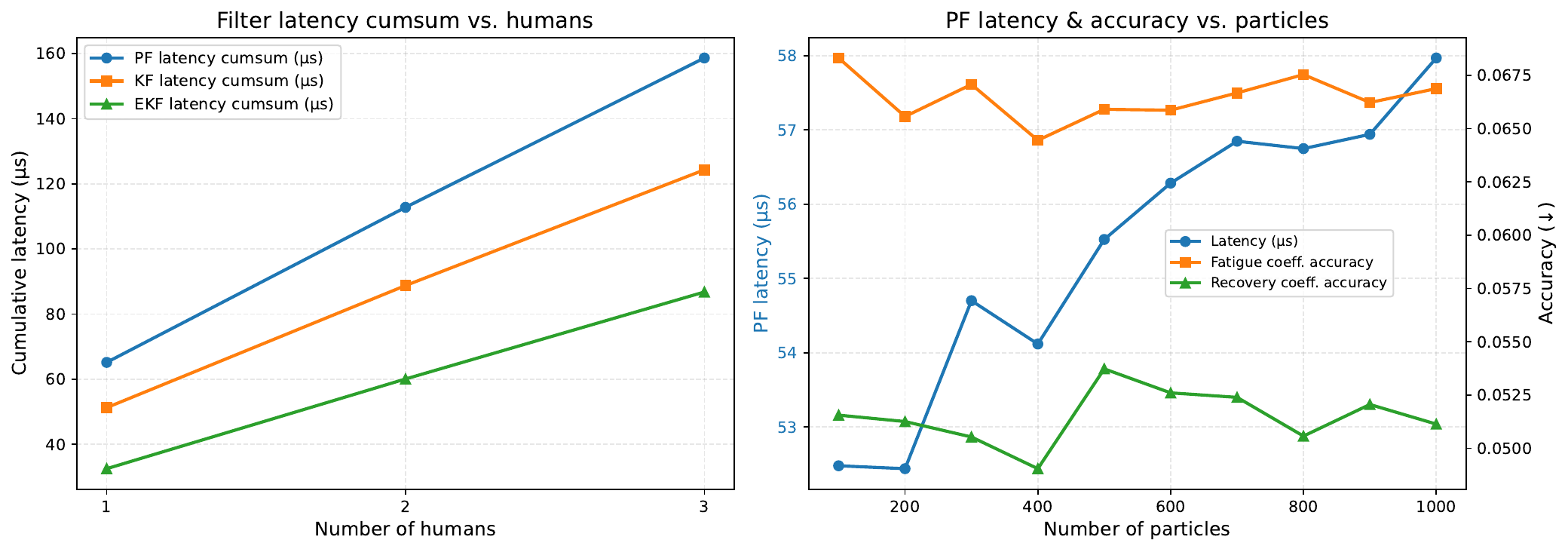}
\caption{Filter latency and algorithm performance in the test stage, varying number particles.\label{fig:filter_time}} 
\end{figure*}

In addition, we examine the PFs' weights update latency. As described in Algorithm \ref{alg:PF_predict} and Sec. \ref{sec:method_pf_predict}, we employ independent PFs for each subtask--human pair, with a default particle count of 500.
Figure \ref{fig:filter_time} further illustrates the latency of particle weight updates as a function of the number of humans and particles. The left subplot shows that, for a single human, the total PF latency across all subtasks is 65 $\mu$s, compared with 51 $\mu$s for the Kalman filter (KF) and 32 $\mu$s for the extended Kalman filter (EKF). As the number of humans increases, the latency exhibits roughly linear growth; for three humans, the PF latency reaches 158 $\mu$s.
The right subplot investigates the effect of particle count (ranging from 100 to 1000). Increasing the number of particles causes only limited fluctuations in the estimation accuracy of the fatigue and recovery parameters, which consistently achieve estimation errors below 0.07 for the fatigue $\lambda$ and below 0.055 for the recovery $\mu$. Meanwhile, the mean latency increases linearly from 52.4 $\mu$s (averaged over 1--3 humans) to 58.0 $\mu$s.

In summary, the task-level fatigue prediction using PF requires only 20 $\mu$s; the complete particle-weight update across all subtasks and three humans takes 158 $\mu$s; and neural-network inference dominates at 2 ms. The total inference time of the entire pipeline comfortably satisfies the real-time requirements of human-robot task planning and allocation in production environments.

\begin{figure*}[b]
\centering	
	\includegraphics[width=0.99 \linewidth, height=0.62\linewidth]{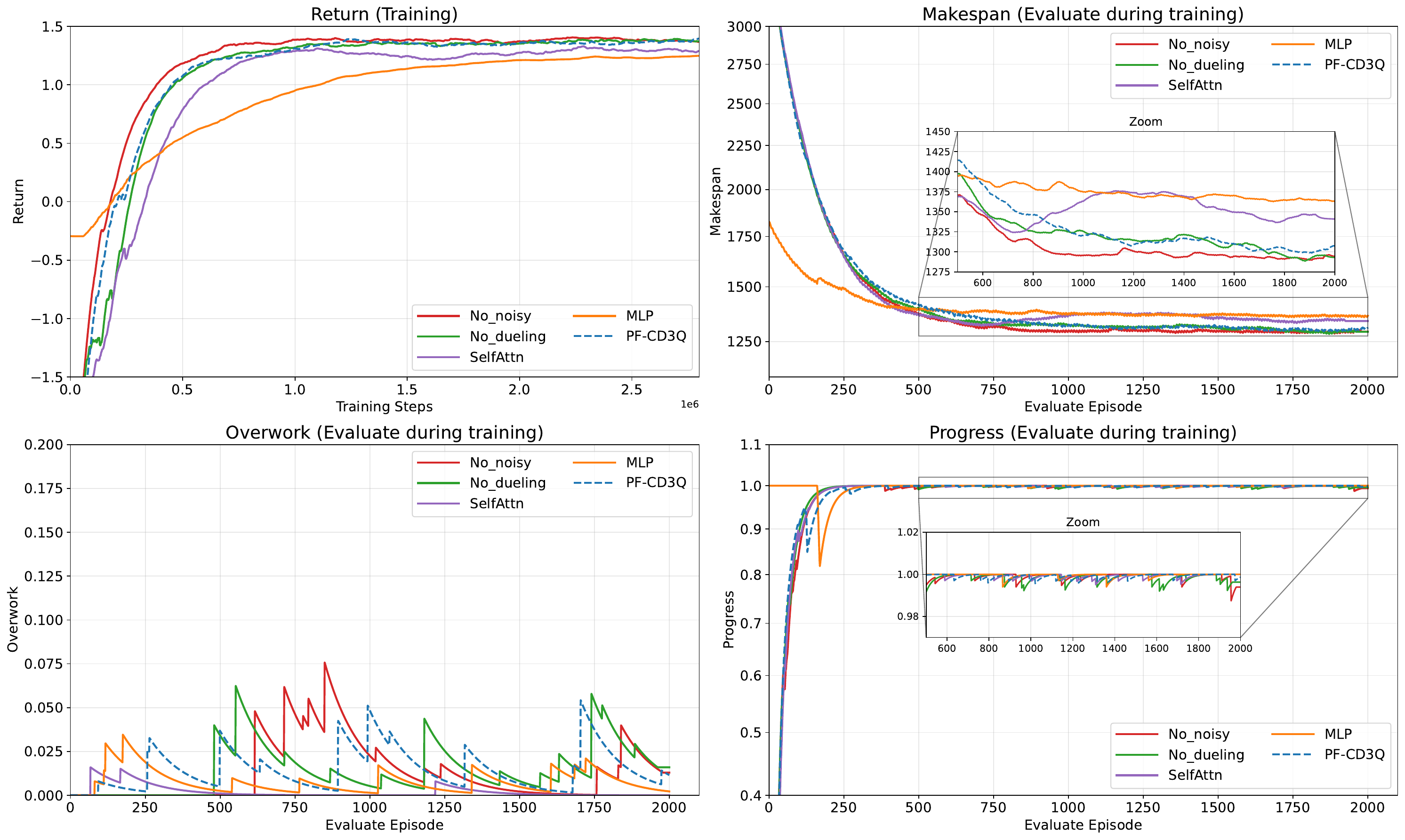}
\caption{Ablation Study: training stage metrics.\label{fig:model_ablation_train}} 
\end{figure*}

\begin{figure*}[b]
\centering	
	\includegraphics[width=0.99 \linewidth, height=0.42\linewidth]{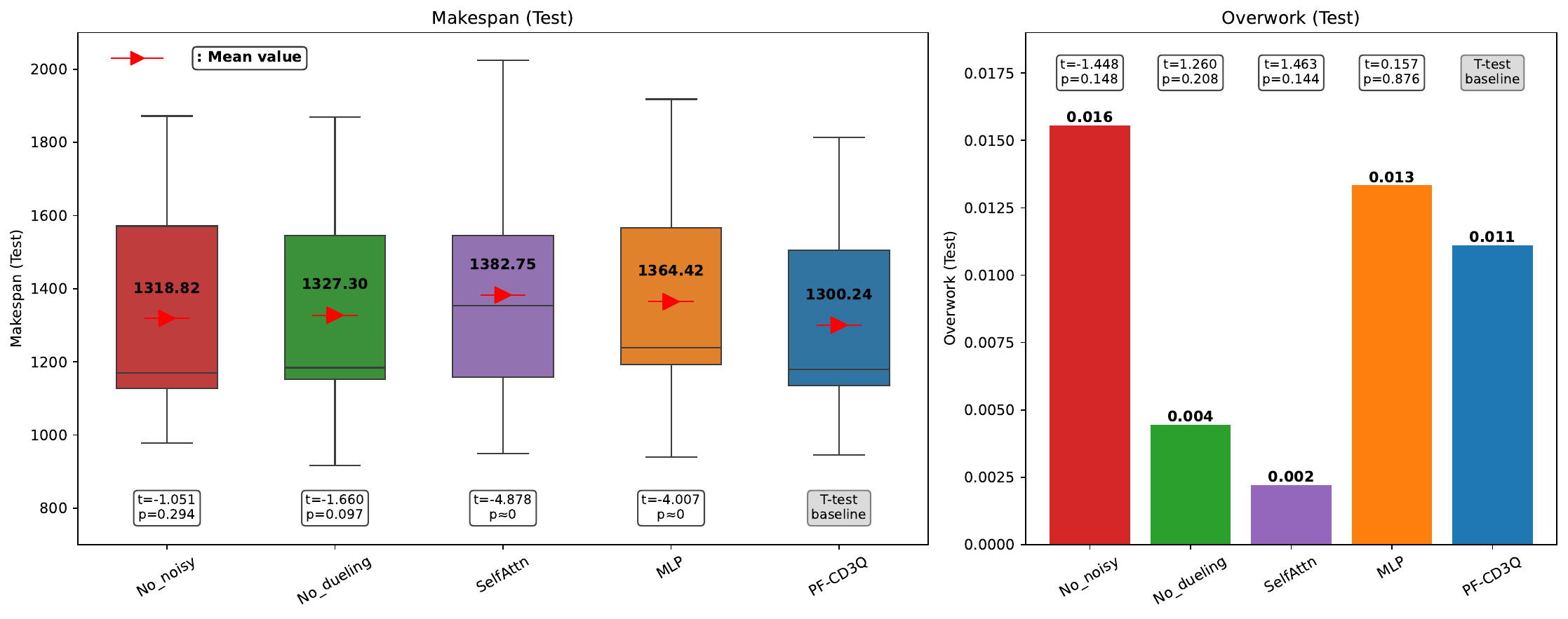}
\caption{Ablation Study: test stage metrics. \label{fig:model_ablation_test}} 
\end{figure*}

\begin{figure*}[b]
\centering	
	\includegraphics[width=0.95 \linewidth, height=0.28    \linewidth]{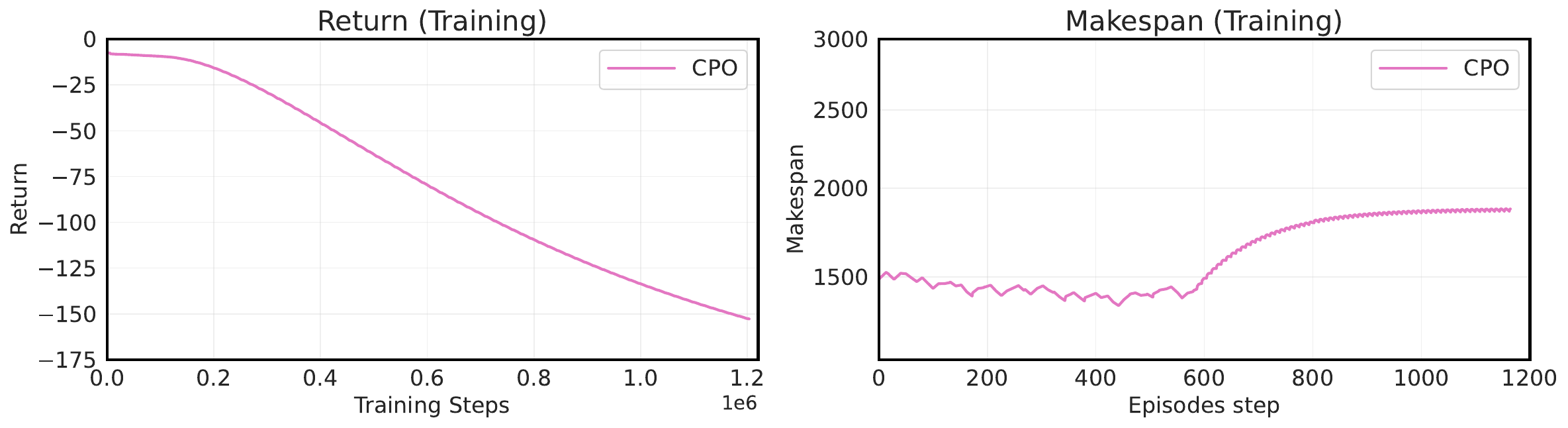}
\caption{The training curves of CPO exhibit clear divergence trends.\label{fig:CPO_training}} 
\end{figure*}

We also conduct additional experiments on model ablations (Fig. \ref{fig:model_ablation_train} and Fig. \ref{fig:model_ablation_test}) and evaluate an additional algorithmic baseline, CPO, in Fig. \ref{fig:CPO_training}. For the model ablation study, we compare our full network architecture (Fig. \ref{fig:network}) against simplified variants, including (i) a pure MLP model, (ii) our architecture without dueling networks \cite{wang2016dueling}, noisy layers \cite{noisynet2017}, or cross-attention (i.e., using only self-attention) \cite{vaswani2017attention}, and (iii) a version without noisy layers. As shown in the training results, the MLP baseline exhibits a clear performance gap in both return and makespan upon convergence, and the self-attention–only variant also shows a noticeable degradation in makespan. In the test stage, PF-CD3Q achieves the best makespan, averaging 1300.24 time steps. The self-attention–only model performs the worst (1382.75), while the no-noisy-layer variant achieves the second-best makespan but suffers from the worst overwork. 
The t-test results further support these differences: for example, both the MLP and the self-attention–only variants show with p-values approaching 0 (larger t-values and smaller p-values indicate stronger statistical differences).
Overall, removing these architectural components leads to varying degrees of performance degradation for PF-CD3Q.

Finally, we also attempted to implement the Constrained Policy Optimization (CPO) algorithm \cite{achiam2017constrained}. Although CPO enforces a trust-region constraint and performs a conjugate-gradient-based second-order optimization step to guarantee theoretically bounded policy improvement, this mechanism is highly sensitive to training noise in non-convex environments. As a result, CPO often becomes unstable and eventually diverges during training, as evidenced in Fig. \ref{fig:CPO_training}.

\clearpage

\begin{table*}[b]
\centering
\small
\renewcommand{\arraystretch}{1.18}
\caption{Training hyper-parameters across each algorithm family.}\label{tab:appendix_train}
\begin{tabularx}{\linewidth}{l *{4}{>{\centering\arraybackslash}X}}
\toprule
Settings &
\makecell{DQN\\PF-DQN} &
\makecell{D3QN\\PF-CD3Q} &
\makecell{PPO\\PF-PPO} &
\makecell{PPO-Lag\\PF-PPO-Lag} \\
\midrule
Network capacity &
\makecell{8.9M} &
\makecell{9.4M} &
\makecell{12.8M} &
\makecell{16.1M} \\
Replay setting &
\makecell{buffer size = 5e5} &
\makecell{buffer size = 5e5} &
\makecell{buffer size = 1e4} &
\makecell{buffer size = 1e4} \\
Optimizer&
\makecell{Adam lr $1\!\times\!10^{-4}$} &
\makecell{Adam lr $1\!\times\!10^{-4}$} &
\makecell{Adam lr $3\!\times\!10^{-4}$} &
\makecell{Adam lr $3\!\times\!10^{-4}$} \\
Warmup steps &
\makecell{5e4} &
\makecell{5e4} &
\makecell{1e3} &
\makecell{1e3} \\
Batch size &
\makecell{512} &
\makecell{512} &
\makecell{512} &
\makecell{512} \\
Q-func update &
\makecell{per 400 steps} &
\makecell{per 400 steps} &
\makecell{nan} &
\makecell{nan} \\
Actor update &
\makecell{nan} &
\makecell{nan} &
\makecell{per 400 steps} &
\makecell{per 400 steps} \\
Critic update &
\makecell{nan} &
\makecell{nan} &
\makecell{per 1000 steps} &
\makecell{per 1000 steps} \\
Cost update &
\makecell{nan} &
\makecell{nan} &
\makecell{nan} &
\makecell{per 1000 steps} \\
Noisynet $\sigma$ &
\makecell{0.1} &
\makecell{0.1} &
\makecell{0.1} &
\makecell{0.1} \\
$\gamma$ &
\makecell{0.99} &
\makecell{0.99} &
\makecell{0.99} &
\makecell{0.99} \\
\bottomrule
\end{tabularx}
\end{table*}

\end{document}